\pdfoutput=1
\documentclass[11pt]{article}

\usepackage[final]{acl}
\usepackage{times}
\usepackage{latexsym}

\usepackage[T1]{fontenc}

\usepackage[utf8]{inputenc}

\usepackage{microtype}

\usepackage{inconsolata}

\usepackage{graphicx}
\usepackage{amsmath}
\usepackage{amssymb}
\usepackage{booktabs}
\usepackage{multirow}
\usepackage{tabularx}
\usepackage[linesnumbered,ruled,vlined]{algorithm2e}
\usepackage{enumitem}
\SetKwInput{KwInput}{Input}
\SetKwInput{KwOutput}{Output}

\title{MIND: An Evidence-State Decision Interface for Criteria-Grounded Psychiatric Inquiry and Diagnosis}

\author{
 \textbf{Guoyi Li\textsuperscript{\textdagger,*,1,2}},
 \textbf{Shihao Xu\textsuperscript{\textdagger,1,3,4}},
 \textbf{Jiatong Ma\textsuperscript{1}},
 \textbf{Zhongjiang Yao\textsuperscript{5}}
\\
 \textbf{Yunyun Han\textsuperscript{1,3}},
 \textbf{Jianhua Chen\textsuperscript{6}},
 \textbf{Yafeng Deng\textsuperscript{1}}
\\
 {\normalfont\normalsize\textsuperscript{1} EverMind AI Inc., California, USA}\\
 {\normalfont\normalsize\textsuperscript{2} Venus Team, Ant Group}\\
 {\normalfont\normalsize\textsuperscript{3} Tianqiao and Chrissy Chen Institute, Shanghai, China}\\
 {\normalfont\normalsize\textsuperscript{4} Lyncia Lab, Shanghai, China}\\
 {\normalfont\normalsize\textsuperscript{5} King's College London, UK}\\
 {\normalfont\normalsize\textsuperscript{6} Shanghai Mental Health Center, Shanghai Jiao Tong University School of Medicine, Shanghai, China}
\\
 {\normalfont\small\href{mailto:zhongjiang.yao@kcl.ac.uk}{zhongjiang.yao@kcl.ac.uk};
 \{guoyi.li,shihao.xu,jiatong.ma,yunyun.han,jianhua.chen,yafeng.deng\}@shanda.com}
}
\begin{document}
\maketitle
\begingroup
\renewcommand{\thefootnote}{\fnsymbol{footnote}}
\footnotetext[2]{Equal contribution.}
\footnotetext[1]{Corresponding author.}
\endgroup
\begin{abstract}
Psychiatric consultation requires agents to elicit discriminative evidence, map uncertain narratives to diagnostic criteria, and decide when evidence suffices. Existing dialogue and retrieval-augmented systems condition policies on raw histories or appended passages, leaving observed evidence, missing checks, differentials, and support reliability entangled. We introduce MIND, a criteria-grounded evidence-state decision interface. At each turn, MIND constructs a typed state containing observed evidence, unresolved criterion checks, active differentials, criterion-linked supports, and reliability metadata. A training-split Psychiatric Reasoning Bank supplies gated supports, turning retrieval into state construction rather than prompt injection. The same state conditions action selection, process rewards, information-gain scoring, and trajectory rectification. Under EMR-grounded simulator protocols, MIND improves accuracy by 8.8 and 7.3 points over strong inference-only, RAG, and RL baselines, with gains transferring to public MDD-5k dialogues. Matched interventions attribute these gains to the shared state rather than prompt length, retrieval text, or formatting. MIND targets screening-level decision support, not autonomous diagnosis. Code is available at \url{https://github.com/Lingxi-mental-health/MIND}.
\end{abstract}

\begin{figure}[t]
\centering
\includegraphics[width=1\columnwidth, height=!]{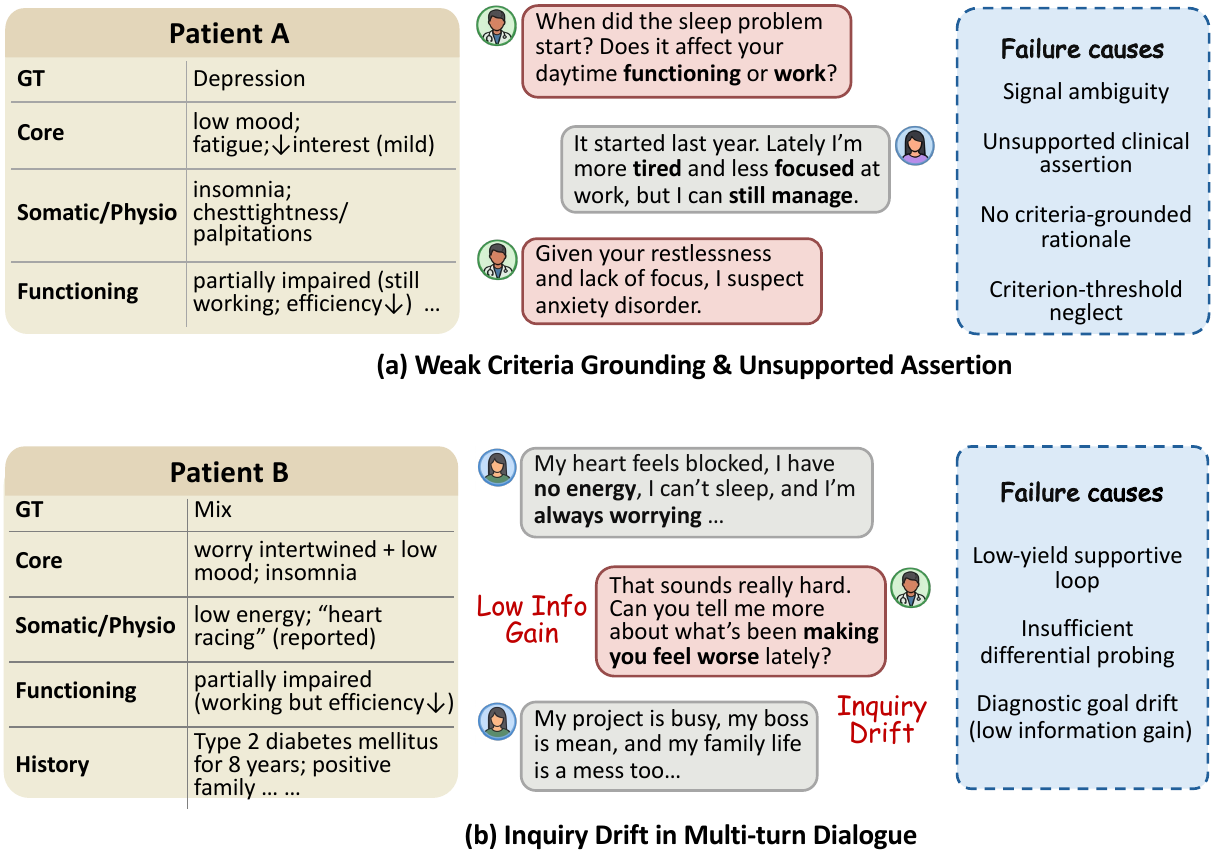}
% \vspace{-2mm}
\caption{Failure modes of implicit clinical state. (a) Without typed evidence, missing checks, and reliability gating, the model makes unsupported assertions. (b) Without tracked missing checks and active differentials, inquiry drifts to generic low-yield questions.}
% \vspace{-5mm}
\label{example1}
\end{figure}

\section{Introduction}

Psychiatric consultation is a demanding form of multi-turn medical dialogue: the doctor must elicit decisive evidence from subjective, incomplete narratives, weigh competing diagnoses against criteria such as DSM-5, and decide when to commit to a screening category \cite{world2022mental,yin2025mdd,feng2026doctoragent}. Unlike symptom-driven dialogue, relevant cues are often metaphorical or temporally fragmented, comorbidity is common, and decisive signals---functional impairment, exclusion checks, safety risks---are easy to miss \cite{lamers2011comorbidity}. An effective agent must therefore track at every turn \emph{what is known}, \emph{what is still missing}, and \emph{which alternative explanations remain live}.

Recent work attacks this problem from several angles: sequential decision methods that reduce diagnostic uncertainty \cite{zhu2025ask,winston2024medical}, retrieval-augmented methods that ground responses in external clinical knowledge \cite{sun2025multi,jeong2024improving}, and joint inquiry--diagnosis frameworks that couple question selection with label prediction \cite{jia2025ddo,dou2025baichuan,aich2025using}. In nearly all of them the policy still conditions on the raw dialogue history $H_t$, optionally augmented with retrieved passages, so which criteria are already satisfied, which are still uncertain, which differentials are active, and how reliable each retrieved support is must be re-derived from text each turn. Two failure modes recur: confident assertions not traceable to the patient report (Figure~\ref{example1}(a)), and inquiry that drifts into generic or repeated questions instead of cues separating competing diagnoses (Figure~\ref{example1}(b)).

\begin{figure}[t]
\centering
\includegraphics[width=1\columnwidth, height=!]{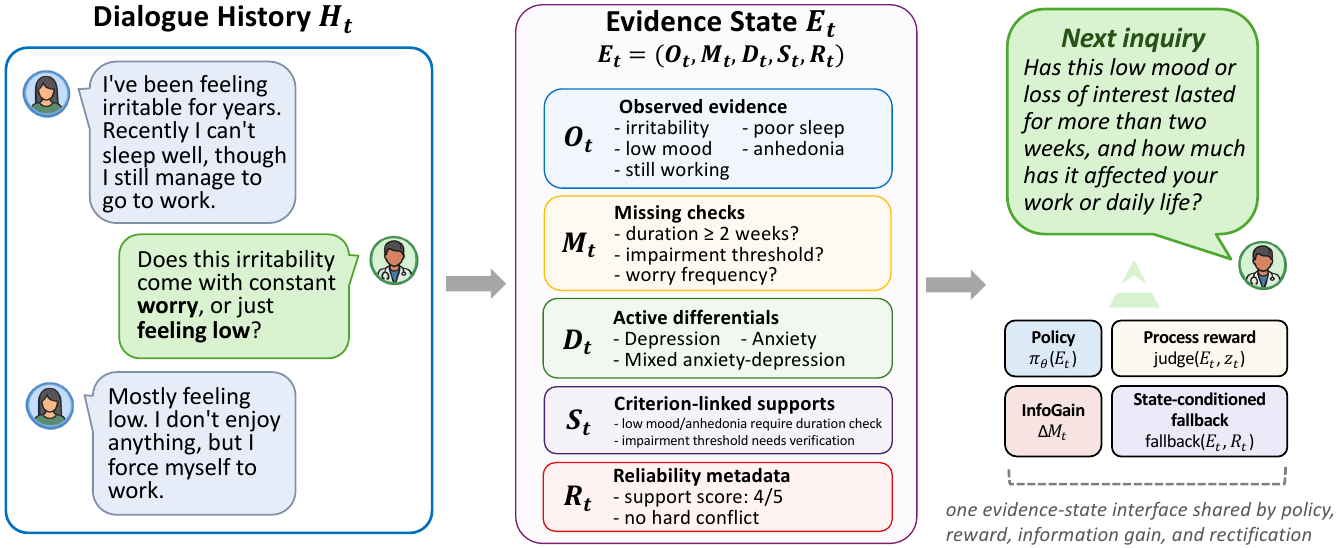}
\caption{MIND in action. (a) Criteria-grounded PRB retrieval drives informative follow-ups. (b) State-conditioned rectification curbs inquiry drift.}
\label{example2}
\end{figure}

We attribute these failures to using $H_t$ as the policy state: the quantities a consultation agent decides over---observed evidence, missing checks, active differentials, and the reliability of external support---are entangled in an unstructured token sequence. We instead introduce an \emph{evidence-state decision interface} (ESDI): a typed, reusable control state that replaces raw dialogue history as the operational decision object. ESDI constructs
\begin{equation}
\begin{aligned}
&\scriptstyle H_t \xrightarrow{f_\phi}
\mathcal{E}_t = (\mathcal{O}_t,\mathcal{M}_t,\mathcal{D}_t,\mathcal{S}_t,\mathcal{R}_t), \\
&\scriptstyle \mathcal{E}_t \rightarrow \{\pi_\theta, R_{\mathrm{proc}}, R_{\mathrm{info}}, \Gamma_{\mathrm{recover}}\},
\end{aligned}
\end{equation}
where $\mathcal{O}_t$ records patient-reported evidence, $\mathcal{M}_t$ unresolved criterion and exclusion checks, $\mathcal{D}_t$ active differential hypotheses, $\mathcal{S}_t$ criterion-linked support entries, and $\mathcal{R}_t$ reliability metadata for those supports. $\mathcal{E}_t$ is not a one-off summary or a prompt template: it is the shared object that the policy, process-reward judge, information-gain scorer, and trajectory-recovery module read and write, so retrieval, criterion checking, reliability assessment, and fallback all act as operators on the same state.

We instantiate this view in MIND, a criteria-grounded reinforcement learning framework for screening-level psychiatric inquiry and diagnostic decision support. MIND builds a Criteria-Grounded Psychiatric Reasoning Bank (PRB) from training-split consultations and clinical criteria, and uses reliability-gated PRB retrieval to populate $\mathcal{S}_t$ and $\mathcal{R}_t$; each retrieved item is admitted only after being mapped to a typed role in $\mathcal{E}_t$, and the admitted fields are reused with identical semantics by the policy, reward judge, information-gain scorer, and fallback. Training combines terminal diagnostic rewards with rubric-based process rewards over $\mathcal{O}_t,\mathcal{M}_t,\mathcal{D}_t$, and a state-conditioned rectification module triggers self-retry or PRB-guided fallback when $\mathcal{R}_t$ or repetition checks indicate a low-yield turn. To our knowledge, this provides a controlled study showing that sharing a typed clinical evidence state across policy learning, reward computation, information-gain scoring, and trajectory recovery improves psychiatric inquiry.

The contributions of this paper are as follows: (1) we introduce an evidence-state decision interface for psychiatric inquiry, replacing raw dialogue history with typed clinical decision variables shared by policy action selection, process rewards, information-gain estimation, and recovery; (2) we operationalize retrieval as state construction rather than prompt augmentation, where PRB supports are admitted only after typed role assignment and reliability gating; (3) we provide controlled empirical evidence that MIND's gains come from the shared typed evidence state rather than prompt formatting or retrieval volume alone, with matched comparisons and representation-level controls showing improvements in diagnostic accuracy, inquiry quality, and support faithfulness.

\begin{figure*}[t]
\centering
\includegraphics[width=2.05\columnwidth, height=!]{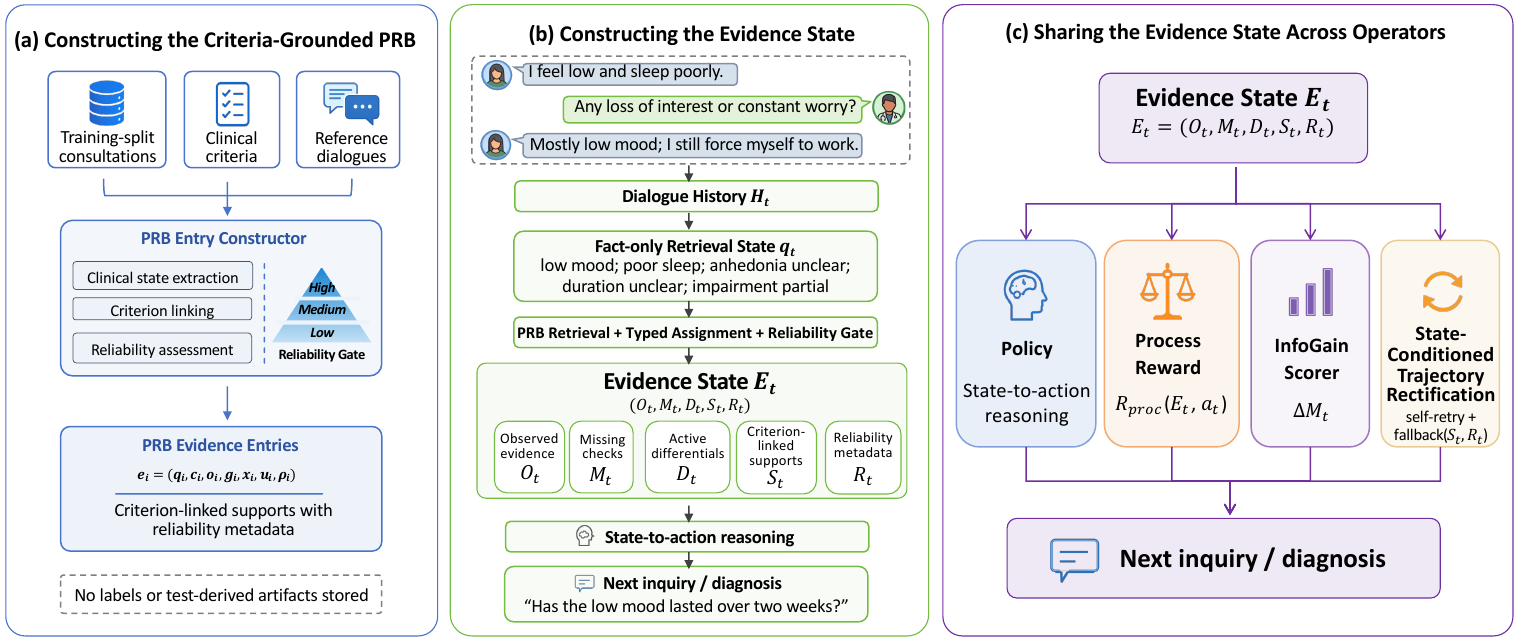}
% \vspace{-2mm}
\caption{MIND architecture. (a) PRB is built offline from training-split consultations as label-free typed entries $e_i=(q_i,c_i,o_i,g_i,x_i,u_i,\rho_i)$. (b) Each turn maps $H_t$ to $\mathcal{E}_t=(\mathcal{O}_t,\mathcal{M}_t,\mathcal{D}_t,\mathcal{S}_t,\mathcal{R}_t)$, with $q_t$ filling observed/missing/differential fields and reliability-gated PRB retrieval filling support/reliability fields. (c) Policy, process reward, information gain, and rectification all read $\mathcal{E}_t$.}
% \vspace{-1mm}
\label{model}
\end{figure*}

\section{Problem Definition}
We formulate psychiatric consultation as a budgeted task-oriented dialogue $\mathcal{P}=\{X^{ex},X^{im},d^\ast\}$, with explicit cues $X^{ex}$, latent cues $X^{im}$ elicited through questioning, and gold label $d^\ast$. At turn $t$, the agent observes history $H_t$, takes an inquiry or diagnosis action $a_t$, and receives a patient response; the episode terminates on a diagnosis or at turn budget $L$, yielding $\tau=(H_1,a_1,\ldots,H_T)$ with $T\le L$ and
\begin{equation}
  \scriptstyle
d_p = \pi_\theta^{\mathrm{diag}}(H_T), \qquad T \le L.
\end{equation}
The task therefore requires both eliciting diagnostically useful evidence within $L$ turns and mapping the accumulated state to $d^\ast$.

\paragraph{Evidence-state MDP formulation.}
Instead of conditioning the policy directly on $H_t$, we treat psychiatric inquiry as an evidence-state MDP in which the operational state is constructed before action selection:
\begin{equation}
\begin{aligned}
  &\scriptstyle H_t \xrightarrow{f_\phi} \mathcal{E}_t, 
  &\scriptstyle a_t \sim \pi_\theta(\mathcal{E}_t,H_t),\quad
   r_t = R(\mathcal{E}_t,a_t,\mathcal{E}_{t+1}).
\end{aligned}
\end{equation}
$\mathcal{E}_t$ is the shared decision interface: the policy reads all typed fields, process rewards score field-consistent reasoning, information gain is defined over the reduction of $\mathcal{M}_t$, and recovery triggers depend on $\mathcal{R}_t$, repetition, and $\mathcal{M}_t$ stagnation. Because individual fields can be masked, shuffled, or gated at inference time, this formulation exposes representation-level intervention points unavailable when the state is raw text.

\section{Methodology}
MIND realizes the evidence-state interface introduced above (Figure~\ref{model}). At turn $t$, a constructor maps $H_t$ to
\begin{equation}
  \scriptstyle
  \mathcal{E}_t =
  (\mathcal{O}_t,\mathcal{M}_t,\mathcal{D}_t,\mathcal{S}_t,\mathcal{R}_t),
\end{equation}
where $\mathcal{O}_t$ holds observed patient facts, $\mathcal{M}_t$ unresolved criterion/exclusion checks, $\mathcal{D}_t$ active differentials, $\mathcal{S}_t$ PRB-retrieved criterion-linked supports, and $\mathcal{R}_t$ their reliability scores and conflict flags. The policy, reward judge, and recovery operator all consume $\mathcal{E}_t$, so the same typed object drives action selection, process supervision, and trajectory rectification. The design has three components: (i) a \emph{Criteria-Grounded Psychiatric Reasoning Bank} (PRB) that populates $(\mathcal{S}_t,\mathcal{R}_t)$ from training-split consultations and clinical criteria; (ii) a retrieval-conditioned policy that performs state-to-action reasoning over $\mathcal{E}_t$; and (iii) SFT$\rightarrow$RL training with terminal rewards, process rewards, and state-conditioned rectification.

\begin{table*}[t]
\centering
\caption{Decision-object comparison: MIND moves the policy/reward/recovery shared object from raw or retrieved text to a typed evidence state $\mathcal{E}_t$. Columns: how evidence enters the loop, the reward variable, and the recovery variable. Each row lists 1--2 anchor works.}
\label{tab:interface_comparison}
\vspace{-1mm}
\resizebox{0.99\textwidth}{!}{
\begin{tabular}{l|l|l|l|l}
\toprule
\textbf{Method type (anchor works)} & \textbf{State object} & \textbf{Retrieval role} & \textbf{Reward object} & \textbf{Recovery object} \\
\midrule
Raw-history RL \cite{feng2026doctoragent,jia2025ddo} & Dialogue history $H_t$ & None & Raw trajectory & None or rules \\
Standard RAG \cite{asai2024self} & $H_t$ plus passages & Prompt context & Raw trajectory & None \\
Structured prompt \cite{chen2024huatuogpt} & Labeled text sections & Formatted context & Raw trajectory & None \\
\textbf{MIND (ours)} & Typed evidence state $\mathcal{E}_t$ & State construction & Evidence-state reward & Evidence-state recovery \\
\bottomrule
\end{tabular}
}
\vspace{-2mm}
\end{table*}

\paragraph{Why a typed state rather than retrieved text.}
$H_t$ plus retrieved passages entangles four control variables---observed evidence, missing checks, active differentials, and retrieval reliability---and forces the policy to re-derive their decomposition as the history grows. Typed factoring makes these variables explicit and keeps $|\mathcal{E}_t|$ bounded while $|H_t|$ grows with $T$. Section~\ref{sec:results_analysis} reports a backbone-, retrieval-, and reward-matched comparison (Table~\ref{tab:state_representation_main}) isolating this interface from prompt formatting and retrieval content.

\paragraph{Evidence-state decision interface.}
A framework satisfies the \emph{evidence-state decision interface} if: (i) it constructs a typed state
\begin{equation}
  \scriptstyle
  \mathcal{E}_t = f_\phi(H_t, \mathrm{PRB})
  = (\mathcal{O}_t,\mathcal{M}_t,\mathcal{D}_t,\mathcal{S}_t,\mathcal{R}_t),
\end{equation}
from history and admissible retrieval; (ii) each field has a distinct decision role; (iii) action selection, reward, information gain, and recovery all read the same state,
\begin{equation}
\begin{aligned}
  &\scriptstyle a_t \sim \pi_\theta(\mathcal{E}_t,H_t),\quad
   r_t=R(\mathcal{E}_t,a_t,\mathcal{E}_{t+1}), \\
  &\scriptstyle \tilde{a}_t=\Gamma_{\mathrm{recover}}(\mathcal{E}_t,a_t);
\end{aligned}
\end{equation}
and (iv) retrieved supports affect decisions only after typed-field assignment and reliability gating. $\mathcal{E}_t$ is therefore an explicit \emph{intervention point}: replacing it with raw history or untyped retrieval under otherwise matched training changes the decision process (Table~\ref{tab:state_representation_main}).

\paragraph{Operational semantics.}
Sharing $\mathcal{E}_t$ across policy, reward, and recovery distinguishes the interface from a structured prompt template, which only organizes text for a single generation call. The \emph{Structured-prompt RAG RL} variant in Table~\ref{tab:state_representation_main} uses the same typed layout in the policy prompt but does not share fields with the reward or fallback. Two medically trained annotators validate $f_\phi$ on 100 turns, reporting $91.8\%$ partial-or-better field accuracy ($77.4\%$ fully correct, Cohen's $\kappa=0.75$).

\paragraph{Complexity argument.}
Since $\mathcal{O}_t,\mathcal{M}_t,\mathcal{D}_t$ live in a fixed schema and $|\mathcal{S}_t|$ is bounded by retrieval depth $k$, $|\mathcal{E}_t|$ stays roughly constant while $|H_t|$ grows linearly in $T$, shrinking the effective hypothesis space on long-horizon dialogues---consistent with the larger PsySim-Adapt drop when PRB, reliability gating, or thinking supervision is removed (Table~\ref{tab:ablation_prb_support}).

\subsection{PRB as a State-Construction Operator}
\label{sec:prb}

PRB is a state-construction operator, not a knowledge bank, answer memory, or generic case retriever: it instantiates the support and reliability fields of $\mathcal{E}_t$ from the current retrieval state. Entries are built offline with an off-the-shelf LLM, but the contribution is the \emph{typed schema and operational contract}: (i) each entry is decomposed into criterion-indexed fields $(c_i,o_i,g_i,x_i,u_i,\rho_i)$ rather than free-form text; (ii) candidates influence the policy only after reliability gating and typed-slot assignment; and (iii) the same entry is reused with role-consistent semantics across action selection, reward, information-gain scoring, and fallback. Swapping the construction LLM leaves downstream performance nearly unchanged (Table~\ref{tab:prb_construction_llm_sensitivity}), indicating the gain comes from the interface contract rather than the extractor.

Formally,
\begin{equation}
  \scriptstyle
  g_{\mathrm{PRB}}(q_t) \;\rightarrow\; (\mathcal{S}_t,\,\mathcal{R}_t),
\end{equation}
and the full evidence state combines $g_{\mathrm{PRB}}$ with fields parsed from $q_t$:
\begin{equation}
  \scriptstyle
  f_\phi(H_t, \mathrm{PRB}) =
  (\mathcal{O}_t,\,\mathcal{M}_t,\,\mathcal{D}_t,\,
  g_{\mathrm{PRB}}^{\mathcal{S}}(q_t),\,
  g_{\mathrm{PRB}}^{\mathcal{R}}(q_t)).
\end{equation}
Each PRB entry is a typed record
\begin{equation}
  \scriptstyle
  e_i=(q_i,c_i,o_i,g_i,x_i,u_i,\rho_i),
\end{equation}
with $q_i$ a compact retrieval state, $c_i$ the associated criterion or exclusion check, $o_i$ the patient-evidence anchor, $g_i$ the missing or unclear check, $x_i$ exclusion/safety cues, $u_i$ a next-inquiry rationale, and $\rho_i$ reliability scores and hard-issue flags. PRB is built only from the training split---no test EMRs, test-derived states, or test support notes---and entries omit diagnostic labels, so it cannot act as an answer memory. Unlike standard RAG, PRB retrieval populates typed fields of $\mathcal{E}_t$ that are reused across inquiry, diagnosis, process reward, and rectification.

\subsubsection{Clinical Retrieval State}
For each training-split dialogue, the pipeline distills a fact-only retrieval state $q_i$ that explicitly marks missing fields as ``not mentioned'' or ``unclear'' and covers chief complaint, duration, severity, sleep, impairment, safety risk, psychosis/mania cues, stressors, and substance use. $q_i$ is then parsed into observed-evidence anchors $o_i$ and missing-check candidates $g_i$, so retrieval updates $\mathcal{O}_t$ and $\mathcal{M}_t$ rather than injecting an undifferentiated paragraph.

\subsubsection{Clinical Support Synthesis}
Given $q_i$, the pipeline synthesizes criterion-indexed fields from the dialogue and clinical references: $c_i$ names the criterion or exclusion check, $x_i$ flags safety/exclusion cues, and $u_i$ records the next-inquiry rationale. These can be rendered as a natural-language note $r_i$, but the policy consumes them through their typed roles in $\mathcal{E}_t$. The LLM acts only as an offline preprocessor; the policy never queries it at inference time.

\subsubsection{Reliability Assessment}
An LLM judge produces $\rho_i$ from dialogue consistency, reference relevance, and any unsupported clinical content, yielding a 1--5 score and hard-issue flags. The raw score is min--max normalized to $[0,1]$ before thresholding. Only entries above the reliability threshold $\tau_{\rm rel}$ enter $\mathcal{S}_t$ and $\mathcal{R}_t$; otherwise MIND falls back to null support.

\subsection{Explicit Clinical Reasoning with Process Supervision}
Each turn proceeds in two retrieval-conditioned steps: build a retrieval state for PRB lookup, then produce support-conditioned state-to-action reasoning under process supervision.

\subsubsection{Step 1: Retrieval-State Construction and PRB Lookup}
Given $H_t$, the policy first verbalizes the observable and missing-check fields of $\mathcal{E}_t$ as a retrieval-oriented query state:
\begin{equation}
  \scriptstyle
y_t^{(1)}=
\texttt{<think>}~z_t^{(1)}~\texttt{</think>}~
\texttt{<rag\_query>}~q_t~\texttt{</rag\_query>}.
\end{equation}
The trace $z_t^{(1)}$ identifies facts, missing fields, and retrieval focus, and is excluded from the core process reward. The resulting $q_t$ approximates $(\mathcal{O}_t,\mathcal{M}_t,\mathcal{D}_t)$ in text and drives PRB lookup. We retrieve the top-$k$ entries by semantic similarity,
\begin{equation}
\begin{aligned}
&\scriptstyle \mathrm{PRB}(q_t)=\mathrm{Top\mbox{-}}k\!\left(\cos(q_t,q_i)\right), \\
&\scriptstyle \widetilde{\mathcal{S}}_t=\{(c_i,o_i,g_i,x_i,u_i)\}_{i=1}^{k},
\end{aligned}
\end{equation}
and only candidates above the reliability threshold form $\mathcal{S}_t$, with their scores and hard-issue flags forming $\mathcal{R}_t$; otherwise the policy uses null support.

\subsubsection{Step 2: State-to-Action Reasoning and Response Generation}
Conditioned on $(H_t,\mathcal{E}_t)$, the policy emits a reasoning trace and the next action:
\begin{equation}
  \scriptstyle
y_t^{(2)}=
\texttt{<think>}~z_t^{(2)}~\texttt{</think>}~
\texttt{<answer>}~a_t~\texttt{</answer>}.
\end{equation}
The trace $z_t^{(2)}$ links observed evidence, missing checks, active alternatives, and reliable supports to the next inquiry or diagnosis.

\subsubsection{Process Reward and RL Objective}
We score the reasoning quality of $z_t^{(2)}$ with an LLM judge along three psychiatric-reasoning dimensions: \emph{symptom analysis} ($S_t^{\mathrm{sym}}$, whether the trace organizes symptoms, onset, severity, and functional impact), \emph{differential consideration} ($S_t^{\mathrm{diff}}$, whether it weighs competing hypotheses and surfaces unresolved criteria), and \emph{decision logic} ($S_t^{\mathrm{dec}}$, whether it justifies the chosen action as informative under the current state). Grounding rewards in these stages avoids generic ``bad reasoning'' penalties and encourages the observation--formulation--decision progression clinicians follow:
\begin{equation}
  \scriptstyle
\mathbf{R}_t^{\mathrm{proc}}=
\big(S_t^{\mathrm{sym}},\,S_t^{\mathrm{diff}},\,S_t^{\mathrm{dec}}\big),
\end{equation}
with each component scored on a 0--2 rubric and normalized before aggregation. The turn-level reward combines $\mathbf{R}_t^{\mathrm{proc}}$ with an information-gain term that credits turns reducing $|\mathcal{M}_t|$. RL then maximizes the episode return
\begin{equation}
  \scriptstyle
J(\theta)=
\mathbb{E}_{\pi_\theta}
\left[
\sum_{t=1}^{T} r_t + r^{\mathrm{term}}
\right],
\end{equation}
where $r_t$ aggregates process reward, information-gain reward, format compliance, and operational penalties, and $r^{\mathrm{term}}$ is the terminal diagnostic reward. Optimization uses GRPO.

\subsubsection{State-Conditioned Trajectory Rectification}
At both training and inference time, MIND monitors each turn for low-utility behaviors---malformed output, repeated inquiries, generic low-yield questions, low retrieval reliability, budget violations, and unsafe content. On a trigger, the agent first attempts a constrained self-retry conditioned on the failure mode; if the retry fails and the fallback cap is not reached, the action is replaced by a structured reference inquiry from the nearest reliable PRB entry. The rule is \emph{state-conditioned} because every trigger is a deterministic function of evidence-state fields (e.g., $|\mathcal{M}_t|$ stagnation, semantic repetition, or low $\mathcal{R}_t$) and requires no separate value network or rollout-level reranking.

\begin{table*}[t]
\centering
\caption{Per-class precision/recall/F1 (\%) on four psychiatric categories under PsySim-Std and PsySim-Adapt. ``$\dagger$'' shares our SFT warm-start data.}
\vspace{-2mm}
\resizebox{\textwidth}{!}{
\begin{tabular}{l|c|l|ccc|ccc|ccc|ccc|c}
\toprule
\multirow{2}{*}{\textbf{PatientSim}} &
\multirow{2}{*}{\textbf{Setting}} &
\multirow{2}{*}{\textbf{Method}} &
\multicolumn{3}{c|}{\textbf{Depression}} &
\multicolumn{3}{c|}{\textbf{Anxiety}} &
\multicolumn{3}{c|}{\textbf{Mix}} &
\multicolumn{3}{c|}{\textbf{Other}} &
\multirow{2}{*}{\textbf{Acc}} \\
\cmidrule(lr){4-6}\cmidrule(lr){7-9}\cmidrule(lr){10-12}\cmidrule(lr){13-15}
 & & & P & R & F1 & P & R & F1 & P & R & F1 & P & R & F1 &  \\
\midrule

% ========================= PsySim-Std =========================
\multirow{15}{*}{\textbf{PsySim-Std}} &
\multirow{8}{*}{Inference-only} &
GPT-4o & 60.0 & 61.0 & 60.5 & 49.5 & 52.0 & 50.7 & 44.0 & 46.0 & 45.0 & 36.0 & 38.0 & 37.0 & 49.5 \\
& & DeepSeek-V3.2 & 49.9 & 42.0 & 45.6 & 30.0 & 38.0 & 33.5 & 48.0 & 44.0 & 45.9 & 60.0 & 58.0 & 59.0 & 46.0 \\
& & Qwen3-4B & 48.0 & 44.5 & 46.2 & 25.0 & 23.0 & 24.0 & 13.5 & 15.0 & 14.2 & 43.5 & 42.0 & 42.7 & 32.8 \\
& & Qwen3-8B & 63.0 & 62.0 & 62.5 & 38.5 & 42.0 & 40.2 & 30.0 & 32.0 & 31.0 & 30.5 & 32.0 & 31.2 & 43.0 \\
& & Qwen3-32B & 63.0 & 62.5 & 62.7 & 38.5 & 36.5 & 37.5 & 32.0 & 34.0 & 33.0 & 51.5 & 49.0 & 50.2 & 47.2 \\
& & Baichuan-M2 & 60.5 & 60.3 & 60.4 & 33.0 & 36.5 & 34.7 & 32.5 & 30.0 & 31.2 & 45.5 & 46.0 & 45.7 & 44.7 \\
& & DDT & 56.0 & 51.0 & 53.4 & 36.0 & 39.0 & 37.4 & 41.0 & 37.0 & 38.9 & 50.0 & 55.0 & 52.4 & 46.5 \\
& & MRD-RAG & 59.0 & 62.0 & 60.5 & 49.0 & 52.0 & 50.5 & 41.0 & 39.0 & 40.0 & 32.0 & 30.0 & 31.0 & 46.0 \\
\cmidrule(lr){2-16}
& \multirow{7}{*}{Fine-tuned} &
Qwen3-4B$^{\dagger}$ 
& 46.0 & 44.0 & 45.0
& 23.5 & 22.0 & 22.7
& 28.0 & 27.5 & 27.7
& 43.0 & 42.5 & 42.7
& 35.5 \\
& & Qwen3-8B$^{\dagger}$ 
& 57.4 & 56.0 & 56.7
& 56.2 & 58.0 & 57.1
& 50.8 & 49.0 & 49.9
& 49.6 & 51.0 & 50.3
& 53.4 \\
& & HuatuoGPT-o1-8B 
& 64.5 & 66.0 & 65.2
& 56.0 & 58.0 & 57.0
& 47.5 & 46.0 & 46.7
& 51.0 & 50.0 & 50.5
& 55.5 \\
& & DoctorAgent-RL
& 65.0 & 64.5 & 64.7
& 61.5 & 63.0 & 62.2
& 55.5 & 54.0 & 54.7
& 58.5 & 59.0 & 58.7
& 60.2 \\
& & DDO
& 65.5 & 64.0 & 64.7
& 61.0 & 62.5 & 61.7
& 56.0 & 54.5 & 55.2
& 59.5 & 58.5 & 59.0
& 60.0 \\
& & Ours (MIND-4B)
& 60.0 & 60.8 & 60.4
& 47.0 & 49.9 & 48.4
& 39.0 & 38.0 & 38.5
& 56.0 & 55.5 & 55.7
& 52.0 \\
& & Ours (MIND-8B) 
& \textbf{75.0} & \textbf{72.5} & \textbf{73.7}
& \textbf{64.5} & \textbf{66.5} & \textbf{65.5}
& \textbf{64.0} & \textbf{63.0} & \textbf{63.5}
& \textbf{73.0} & \textbf{72.0} & \textbf{72.5}
& \textbf{69.0} \\
\midrule

% ========================= PsySim-Adapt =========================
\multirow{15}{*}{\textbf{PsySim-Adapt}} &
\multirow{8}{*}{Inference-only} &
GPT-4o & 54.0 & 53.0 & 53.5 & 40.0 & 42.0 & 41.0 & 36.0 & 38.0 & 37.0 & 27.0 & 28.0 & 27.5 & 40.5 \\
& & DeepSeek-V3.2 & 40.0 & 33.0 & 36.2 & 22.1 & 26.5 & 24.1 & 42.1 & 37.0 & 39.4 & 50.0 & 48.0 & 49.0 & 36.8 \\
& & Qwen3-4B & 30.0 & 29.5 & 29.7 & 18.0 & 18.0 & 18.0 & 22.0 & 22.0 & 22.0 & 25.0 & 24.5 & 24.7 & 24.2 \\
& & Qwen3-8B & 45.0 & 44.0 & 44.5 & 30.0 & 30.0 & 30.0 & 24.0 & 24.0 & 24.0 & 33.0 & 32.0 & 32.5 & 33.4 \\
& & Qwen3-32B & 51.0 & 50.0 & 50.5 & 31.0 & 29.0 & 30.0 & 24.0 & 24.0 & 24.0 & 42.0 & 42.0 & 42.0 & 37.8 \\
& & Baichuan-M2 & 50.0 & 48.5 & 49.2 & 28.0 & 27.0 & 27.5 & 22.0 & 22.0 & 22.0 & 41.0 & 40.0 & 40.5 & 35.9 \\
& & DDT & 45.0 & 44.0 & 44.5 & 31.0 & 29.0 & 30.0 & 33.0 & 28.0 & 30.3 & 47.0 & 43.0 & 44.9 & 37.1 \\
& & MRD-RAG & 50.4 & 51.6 & 51.0 & 39.5 & 46.6 & 42.8 & 33.0 & 27.3 & 29.9 & 22.4 & 20.5 & 21.4 & 36.5 \\
\cmidrule(lr){2-16}

& \multirow{7}{*}{Fine-tuned}&
Qwen3-4B$^{\dagger}$ 
& 39.0 & 36.5 & 37.7
& 12.0 & 11.0 & 11.5
& 23.0 & 22.0 & 22.5
& 33.0 & 32.0 & 32.5
& 27.0 \\
& & Qwen3-8B$^{\dagger}$
& 48.0 & 46.0 & 47.0
& 43.0 & 42.0 & 42.5
& 39.0 & 38.0 & 38.5
& 47.0 & 48.0 & 47.5
& 44.0 \\
& & HuatuoGPT-o1-8B 
& 58.7 & 59.4 & 59.0 
& 51.0 & 49.9 & 50.4 
& 41.5 & 35.4 & 38.2 
& 30.4 & 28.5 & 29.4 
& 43.5 \\
& & DoctorAgent-RL
& 59.0 & 57.5 & 58.2
& \textbf{54.0} & \textbf{52.5} & \textbf{53.2}
& 42.0 & 43.0 & 42.5
& 50.0 & 55.0 & 52.4
& 52.5 \\
& & DDO
& 55.0 & 53.5 & 54.2
& 49.0 & 52.0 & 50.5
& \textbf{47.0} & 45.0 & 46.0
& 50.0 & 47.0 & 48.5
& 49.4 \\
& & Ours (MIND-4B)
& 50.0 & 48.0 & 49.0
& 47.0 & 46.0 & 46.5
& 41.0 & 39.0 & 40.0
& 47.0 & 46.0 & 46.5
& 45.0 \\
& & Ours (MIND-8B)
& \textbf{69.0} & \textbf{67.0} & \textbf{68.0}
& 53.0 & 51.0 & 52.0
& \textbf{47.0} & \textbf{46.0} & \textbf{46.5}
& \textbf{71.0} & \textbf{70.0} & \textbf{70.5}
& \textbf{59.8} \\
\bottomrule
\end{tabular}
}
\vspace{-2mm}
\label{tab:perclass_prf_patientsim}
\end{table*}

\section{Experiments}

\textbf{Datasets.}
We use 1{,}600 de-identified Chinese psychiatric EMRs (2019--2023) over four ICD-10-style screening categories~\cite{world1992icd} (Depression, Anxiety, Mixed, Other; adjudicated $\kappa=0.78$). EMR-level splits are taken once before any artifact is produced: 1{,}200 training EMRs feed PRB construction, SFT, and GRPO; 400 held-out EMRs drive PsySim-Std/PsySim-Adapt closed-loop evaluation. No test EMR, test-derived profile, simulated dialogue, or support note enters the PRB or any training stage. Statistics, ethics, release, and the public MDD-5k set~\cite{yin2025mdd} are in Appendix~\ref{app_es},~\ref{app:dataset_release}; code and a de-identified sample: \url{https://anonymous.4open.science/r/MIND-83DF/}.

\textbf{Baselines.}
We compare inference-only LLMs (GPT-4o, DeepSeek-V3.2, Qwen3-4B/8B/32B, Baichuan-M2), retrieval/criteria-guided systems (DDT, MRD-RAG), and fine-tuned/RL consultation models (Qwen3-4B/8B$^{\dagger}$, HuatuoGPT-o1-8B, DoctorAgent-RL, DDO). $\dagger$ marks the same SFT warm-start data as MIND. Implementation and fairness controls are in Appendix~\ref{app:baselines_impl},~\ref{app:baseline_fairness}.

\textbf{Experimental Setup.}
MIND uses Qwen3 backbones with a PRB-derived SFT warm start and GRPO, optimizing terminal correctness, process supervision, information gain, retrieval reliability, and operational penalties. All systems share a turn budget ($L\!=\!10$), final-diagnosis protocol, and four-way label space, and are evaluated under two EMR-grounded simulators (PsySim-Std, PsySim-Adapt) plus the simulator-independent public MDD-5k. Gold-label accuracy and macro-F1 are the primary effectiveness metrics; support-faithfulness and process-quality judges are scalable analysis instruments rather than clinical validators, and human ratings provide a complementary assessment. We report 95\% paired bootstrap CIs against the strongest baseline; full hyperparameters and statistical protocols are in Appendix~\ref{app_es}.

% \paragraph{Additional experiments in the appendix.}
% To keep the main paper focused, several controlled studies and robustness checks are reported in the appendix and cited inline below: simulator audit and cross-backbone simulators (Appendix~\ref{app:cross_backbone_simulator}); PRB leakage audit (Appendix~\ref{app:prb_leakage_audit}); evidence-state construction quality (Appendix~\ref{app:state_construction_quality}); field masking (Appendix~\ref{app:field_masking}); prompt-, token-, and retrieval-level controls (Appendix~\ref{app:prompt_retrieval_controls}); PRB construction-LLM sensitivity (Appendix~\ref{app:prb_construction_llm_sensitivity}); judge robustness and human--judge correlation (Appendix~\ref{app:judge_robustness}, Appendix~\ref{app:human_judge_correlation}); MCR protocol (Appendix~\ref{app:mcr_details}); external MDD-5k diagnosis and prefix next-question (Appendix~\ref{app:external_mdd5k_eval}, Appendix~\ref{app:mdd5k_prefix_next_question}); fine-grained analyses and parameter study (Appendix~\ref{app:fine_grained_analyses}); Other-category composition (Appendix~\ref{app:other_composition}); diagnostic--dialogue balance visualization (Appendix~\ref{app:radar_balance}); and safety evaluation (Appendix~\ref{app:safety_evaluation}).

\begin{table}[t]
\centering
\caption{Public MDD-5k four-way diagnosis from released dialogue history. Set is excluded from PRB, SFT, RL, reward, simulator, and hyperparameter tuning.}
\label{tab:mdd5k_main_diag}
\vspace{-1mm}
\resizebox{0.53\columnwidth}{!}{
\begin{tabular}{lcc}
\toprule
Model & Acc & Macro-F1 \\
\midrule
Qwen3-8B & 34.8 & 32.9 \\
Qwen3-8B$^{\dagger}$ & 42.5 & 40.6 \\
MRD-RAG & 38.6 & 36.8 \\
DoctorAgent-RL & 46.8 & 45.0 \\
DDO & 48.6 & 47.0 \\
MIND-8B & \textbf{53.8} & \textbf{52.0} \\
\bottomrule
\end{tabular}
}
\vspace{-5mm}
\end{table}

\textbf{Patient Simulator Validation.}
\label{sec:simulator_validation}
Since simulator and policy may share a Qwen3~\cite{yang2025qwen3} backbone, we audit the simulator first. Profiles exclude gold labels and forbid out-of-profile symptoms; an automatic audit covers profile factual consistency, unsupported symptom rate, label leakage, and answerability, and two medically trained annotators validate 100 responses (Cohen's $\kappa{=}0.81$). Label leakage is zero by construction and factual consistency exceeds $90\%$ under both simulators (Table~\ref{tab:simulator_validation}). Re-running under GLM~4.7~\cite{5team2025glm45agenticreasoningcoding} and Llama-3-8B-Instruct~\cite{grattafiori2024llama} simulators keeps the MIND-over-DDO gap at $+8.5$ to $+8.9$ (Appendix~\ref{app:cross_backbone_simulator}), and non-matching simulator--policy pairings preserve the ordering (Appendix~\ref{app:simulator_policy_pairing}); full prompts and protocol are in Appendix~\ref{app_es}.

\begin{table}[t]
\centering
\caption{Patient simulator audit: profile factual consistency, unsupported symptom rate, label leakage, and answerability over all simulator turns.}
\label{tab:simulator_validation}
\vspace{-1mm}
\resizebox{0.48\textwidth}{!}{
\begin{tabular}{lcc}
\toprule
\textbf{Metric} & \textbf{PsySim-Std} & \textbf{PsySim-Adapt} \\
\midrule
Profile factual consistency (\%) $\uparrow$ & 94.2 & 91.6 \\
Unsupported symptom rate (\%) $\downarrow$  & 3.1  & 5.4  \\
Label leakage rate (\%) $\downarrow$        & 0.0  & 0.0  \\
Answerability rate (\%) $\uparrow$          & 92.7 & 88.9 \\
\bottomrule
\end{tabular}
}
\vspace{-2mm}
\end{table}

\subsection{Results and Analysis}
\label{sec:results_analysis}
\textbf{Overall Performance.}
Table~\ref{tab:perclass_prf_patientsim} reports per-class P/R/F1 under both simulators. MIND achieves the best overall accuracy and macro-F1 under both simulators, with category-level gains largest on Depression and Other and more mixed on Anxiety and Mix. The MIND-over-DDO gap is largest when early narratives admit multiple labels and the agent must judge which criterion is unresolved, when to stop, and which retrieved cue is reliable, indicating that criteria-grounded PRB retrieval, reliability gating, and process-supervised state-to-action reasoning---rather than turn-budget effects---drive the gain.

\textbf{State-Representation Comparison.}
All variants share the Qwen3-8B backbone, SFT data, GRPO setup, reward weights, retrieval content, and turn budget; the policy and reward state-sharing configuration differs (Table~\ref{tab:state_representation_main}). Raw history and standard RAG trail MIND, prompt-only structuring yields a modest gain, one-sided sharing of the typed state is intermediate, and the largest jump occurs when policy and reward both operate on the evidence state---identifying \emph{policy--reward consistency over a shared typed state}, rather than retrieval content or prompt layout, as the main source of improvement.

\begin{table}[t]
\centering
\caption{Matched-variable controls over policy/reward state sharing (Qwen3-8B, same SFT/GRPO/rewards/retrieval/turn budget). The six control rows disable fallback to isolate representation; two one-sided rows share the typed state with only policy or only reward to isolate policy--reward consistency.}
\label{tab:state_representation_main}
\vspace{-1mm}
\resizebox{0.48\textwidth}{!}{
\begin{tabular}{lcccc}
\toprule
\multirow{2}{*}{\textbf{Variant}} &
\multicolumn{2}{c}{\textbf{PsySim-Std}} &
\multicolumn{2}{c}{\textbf{PsySim-Adapt}} \\
\cmidrule(lr){2-3}\cmidrule(lr){4-5}
& \textbf{Acc} & \textbf{Macro-F1} & \textbf{Acc} & \textbf{Macro-F1} \\
\midrule
Raw-history RL & 63.4 & 64.5 & 53.9 & 54.4 \\
Standard RAG RL & 66.0 & 65.8 & 56.5 & 56.9 \\
Structured-prompt RAG RL & 66.7 & 66.5 & 57.2 & 57.5 \\
Policy=ES, Reward=Raw    & 67.2 & 67.2 & 57.8 & 58.0 \\
Policy=Raw, Reward=ES    & 66.4 & 66.0 & 56.9 & 57.0 \\
Evidence-state RL (both) & 68.2 & 67.5 & 59.0 & 58.5 \\
\midrule
MIND & \textbf{69.0} & \textbf{68.8} & \textbf{59.8} & \textbf{59.3} \\
\bottomrule
\end{tabular}
}
\vspace{-2mm}
\end{table}

\textbf{Public MDD-5k Diagnostic Evaluation.}
On released MDD-5k~\cite{yin2025mdd} dialogues, with no follow-up and a fixed history, MIND raises accuracy from 48.6 to 53.8 and macro-F1 from 47.0 to 52.0 over the strongest baseline (Table~\ref{tab:mdd5k_main_diag}). This is a simulator-independent transfer test, not a substitute for closed-loop active inquiry; the complementary prefix task evaluates next-question selection without simulated responses. Protocol and prefix-based results are in Appendix~\ref{app:external_mdd5k_eval},~\ref{app:mdd5k_prefix_next_question}.

\begin{table}[t]
\centering
\caption{MIND component ablation under two simulators. Support is the FC/SG/PF average from the support-faithfulness rubric. Generalization of PRB to non-MIND backbones (Qwen3-32B / Baichuan-M2 $\pm$ PRB) is in Appendix~\ref{app:prb_external_backbones}.}
\label{tab:ablation_prb_support}
\vspace{-1mm}
\resizebox{0.48\textwidth}{!}{
\begin{tabular}{l|ccc|ccc}
\toprule
\multirow{2}{*}{\textbf{Model}} &
\multicolumn{3}{c|}{\textbf{PsySim-Std}} &
\multicolumn{3}{c}{\textbf{PsySim-Adapt}} \\
\cmidrule(lr){2-4}\cmidrule(lr){5-7}
& \textbf{Support} & \textbf{Acc} & \textbf{F1} & \textbf{Support} & \textbf{Acc} & \textbf{F1} \\
\midrule

MIND w/o SFT               & \large{8.0} & \large{62.7} & \large{64.0} & \large{7.8} & \large{53.7} & \large{54.5} \\
MIND w/o InfoGain reward   & \large{8.2} & \large{66.5} & \large{67.7} & \large{8.0} & \large{57.5} & \large{57.8} \\
MIND w/o Reasoning reward  & \large{8.1} & \large{65.3} & \large{66.9} & \large{7.9} & \large{56.0} & \large{56.4} \\
\midrule

MIND w/o PRB               & \large{8.0} & \large{65.7} & \large{66.1} & \large{7.7} & \large{56.3} & \large{57.0} \\
MIND w/o reliability gate  & \large{8.1} & \large{66.5} & \large{67.2} & \large{7.8} & \large{57.1} & \large{57.7} \\
MIND w/o thinking          & \large{7.5} & \large{61.5} & \large{60.6} & \large{7.0} & \large{51.9} & \large{52.3} \\
MIND w/o fallback          & \large{8.2} & \large{67.1} & \large{67.8} & \large{8.1} & \large{58.0} & \large{58.5} \\
\midrule

\textbf{MIND} &
\large{\textbf{8.3}} & \large{\textbf{69.0}} & \large{\textbf{68.8}} &
\large{\textbf{8.3}} & \large{\textbf{59.8}} & \large{\textbf{59.3}} \\
\bottomrule
\end{tabular}
}
\vspace{-2mm}
\end{table}

\textbf{Ablation Study.}
Table~\ref{tab:ablation_prb_support} decomposes MIND into training, retrieval, reasoning, and recovery. Across both simulators, removing SFT weakens the two-step turn format, dropping reasoning or information-gain rewards reduces evidence acquisition, and removing PRB or reliability gating lowers both support quality and accuracy; disabling fallback has a smaller but stable effect, matching its role as low-yield-turn repair rather than a representation change. PRB also transfers to non-MIND backbones (Qwen3-32B, Baichuan-M2) under both simulators (Appendix~\ref{app:prb_external_backbones}). Field masking, token-matched RAG, random retrieval, and shuffled reliability further show $\mathcal{E}_t$ is not a prompt template (Appendix~\ref{app:field_masking},~\ref{app:prompt_retrieval_controls}); matched state-abstraction, separated-judge, scale-behavior, and paired-significance results are in Appendix~\ref{app:belief_slot_comparison},~\ref{app:judge_family_separation},~\ref{app:backbone_scale_behavior},~\ref{app:control_significance}.

\begin{figure}[t]
\centering
\includegraphics[width=0.9\columnwidth]{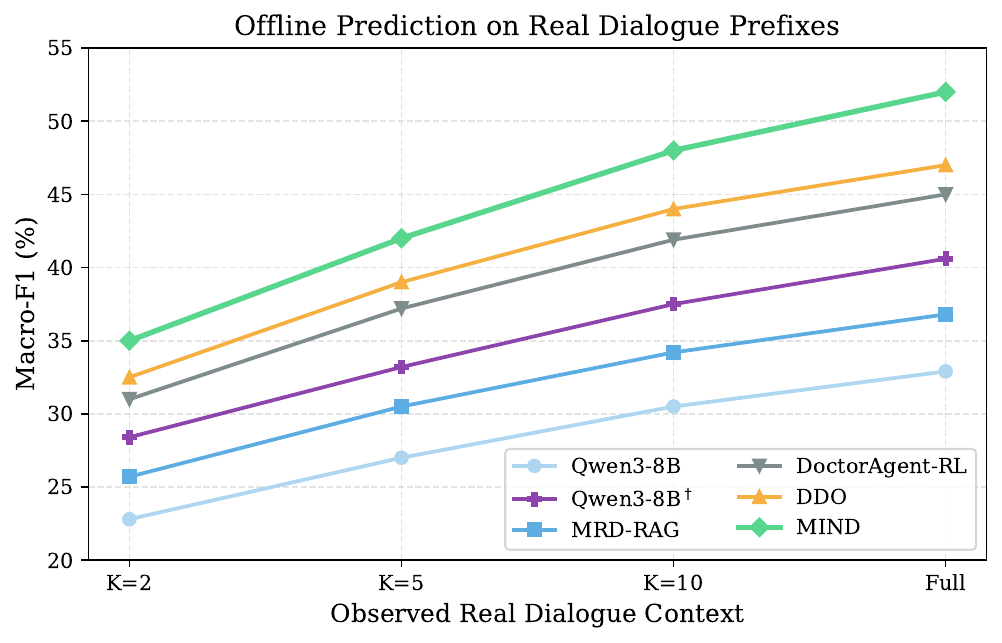}
\vspace{-2mm}
\caption{Offline screening on public MDD-5k prefixes: MIND generally scales best with longer public histories, suggesting evidence-state aggregation transfers outside simulator rollout.}
\label{fig:offline_real_prefix_curve}
\end{figure}

\textbf{Offline Scaling on Public Prefixes.}
On increasing prefix lengths from public MDD-5k dialogues (Figure~\ref{fig:offline_real_prefix_curve}), MIND's macro-F1 generally improves and its gap over the strongest baseline is maintained at longer prefixes, showing that the evidence-state policy converts longer public histories into better screening predictions. Reliability-threshold $\tau_{\rm rel}$ sensitivity (gold-label peak at $0.2$) and an independent GPT-4o cross-judge for faithfulness---separate from the LLM producing $\rho_i$---are in Appendix~\ref{app:fine_grained_analyses},~\ref{app:judge_robustness}.

\textbf{Inquiry-Quality Indicator.}
MIND attains the highest Missing-Check Resolution (MCR-shared $49.6$ vs.\ DDO $41.2$) and the lowest repetition ($8.0$ vs.\ $12.1$), reflecting more targeted evidence acquisition (Appendix~\ref{app:mcr_details}).

\textbf{Support Faithfulness Evaluation.}
Each dialogue is paired with PRB-retrieved turn-level supports; DeepSeek-V3.2 scores anonymized outputs on factual consistency (FC), support grounding (SG), and patient faithfulness (PF). MIND attains the top average (8.3), $+0.8$ over DDO and $2.0$--$2.5$ over inference-only and RAG baselines, with the largest gap on SG ($+0.8$), consistent with support-conditioned decisions (Appendix Table~\ref{tab:support_faithfulness_eval}). An independent GPT-4o judge preserves the ranking (Spearman $0.93$); human--judge correlations are in Appendix~\ref{app:judge_robustness},~\ref{app:human_judge_correlation}.

\begin{table}[t]
\centering
\caption{Human evaluation on 160 cases (40 per category) by five reviewers (one licensed psychiatrist, four medically trained graduate students).}
\label{tab:human_eval}
\vspace{-1mm}
\resizebox{0.48\textwidth}{!}{
\begin{tabular}{lccc}
\toprule
Model & Professional Reliability   & Naturalness   & Preference (\%)  \\
\midrule
GPT-4o        & 2.95 & 3.32 & 45.0 \\
Qwen3-8B      & 2.15 & 2.80 & 22.5 \\
Baichuan-M2   & 2.50 & 3.05 & 32.0 \\
DDO           & 3.15 & 3.20 & 52.5 \\
Ours (MIND)   & \textbf{4.05} & \textbf{3.92} & \textbf{78.5} \\
\bottomrule
\end{tabular}
}
\vspace{-5mm}
\end{table}

\textbf{Human Evaluation.}
Five reviewers blindly rate 160 anonymized cases on Professional Reliability, Naturalness, and pairwise Preference, with large PR disagreements adjudicated by the psychiatrist. Preference (\%) is the fraction of pairwise comparisons in which a majority of reviewers preferred the model's output over each of the other four systems; since each model is compared against four opponents, values across models need not sum to 100\%. MIND leads on PR (4.05), Naturalness (3.92), and preference (78.5\%), with pairwise agreement 0.79 (inter-reviewer agreement on pairwise preference decisions) and pre-adjudication Krippendorff's $\alpha=0.74$ (Table~\ref{tab:human_eval}); a balanced diagnostic--dialogue radar is given in Appendix~\ref{app:radar_balance}.


\section{Related Work}

\subsection{Psychiatric Consultation as Multi-turn Inquiry}
Multi-turn medical consultation has been cast as sequence generation, follow-up selection, and uncertainty-reducing planning over patient narratives \cite{valizadeh2022ai,tang2016inquire,wei2018task,kao2018context,chen2022diaformer,wang2023coad,hou2023mtdiag,zhao2024haiformer,hu2024uncertainty,chopra2025feedback,chen2025cod,xu2026lingxidiagbench}. Psychiatric consultation differs in that diagnostic thresholds, exclusion checks, functional impairment, and safety risks are seldom explicit, causing drift to low-yield questions or unsupported assertions \cite{xia2020generative,xu2019end,liu2022my,zhong2022hierarchical,yin2025mdd}; MIND addresses this with criteria-grounded policy learning.

\subsection{Medical LLMs for Sequential Decision-Making}
Medical LLMs have advanced history taking, diagnosis, and treatment planning via prompting, domain adaptation, CoT, and instruction tuning \cite{zhou2025large,kim2024mdagents,shi2024medadapter,johri2024craft,jia2025medikal,rose2025meddxagent,zheng2024exploring,liu2024medcot,tian2024chimed,wang2025citrus}, with recent moves toward proactive consultation, multi-agent collaboration, and RL-optimized inquiry \cite{tu2025towards,ma2025memorize,zhang2024cpsycoun,liu2023chatcounselor,jia2025ddo,dou2025baichuan,pan2025medvlm,lai2025med,shao2024deepseekmath,chen2024huatuogpt,feng2026doctoragent}. In these systems policy, reward, and recovery typically read raw history or appended retrieval independently, yielding sparse long-horizon supervision. Unlike generic process-reward RL, MIND changes both \emph{what is optimized} and \emph{what is acted upon}: policy, reward, information gain, and recovery consume the same typed state.

\subsection{Retrieval-Augmented Medical Reasoning}
RAG grounds medical LLMs in guidelines, references, or case knowledge \cite{jeong2024improving,asai2024self,li2023meddm}, but unfiltered passages can mislead inquiry in metaphorical or underspecified psychiatric narratives. MIND turns retrieval into state construction: a clinical retrieval state queries the Psychiatric Reasoning Bank, and admitted entries populate typed fields shared by policy, reward, and recovery \cite{yin2025mdd,tu2025towards}. Structured prompting can organize the same content for one generation call, but it does not by itself define a state shared with reward and fallback. Likewise, POMDP belief states \cite{young2013pomdp,williams2007partially} and slot-based DST \cite{wu2019transferable} primarily track category beliefs or task slots for dialogue management; $\mathcal{E}_t$ instead carries observed evidence, unresolved checks, active differentials, criterion-linked supports, and reliability metadata as a criteria-grounded decision interface shared across system components.

\section{Conclusion}
MIND is a criteria-grounded RL framework that recasts screening-level psychiatric inquiry as evidence-state decision making. A training-split Psychiatric Reasoning Bank turns retrieval into a shared state for policy learning, process rewards, information-gain scoring, and trajectory recovery. Across EMR-grounded simulators and MDD-5k, MIND improves accuracy, support faithfulness, and inquiry quality over inference-only, RAG, and RL baselines. Matched-variable interventions further show that these gains come from the shared evidence-state interface rather than prompt format or retrieval volume, supporting evidence-state design for long-horizon criteria-grounded medical dialogue.

\section*{Limitations}
This work has several limitations. First, we study a four-way psychiatric screening setup rather than the full granularity of clinical coding. The labels and reported decisions should therefore be interpreted as screening-level decision support, not full psychiatric diagnosis; the heterogeneous composition of Other is summarized in Appendix~\ref{app:other_composition}. Symptom overlap, comorbidity, subjective reporting, and temporal variation remain difficult, particularly near the Anxiety--Mixed boundary.

Second, evidence-state construction is imperfect. The 100-turn annotation is a pilot-scale consistency audit rather than clinical-grade validation, and field masking does not replace systematic item-level error injection into symptoms, missing checks, differentials, supports, and reliability metadata. Such perturbation analysis, together with non-LLM reliability estimation and robustness to biased process rewards, remains future work. LLM judges are used as scalable analysis instruments, not clinical validators.

Third, the main interactive evaluation remains simulator-based. Public MDD-5k fixed-history and prefix evaluations reduce simulator dependence but do not replace prospective clinician-in-the-loop interaction. Calibrated stopping, premature-closure analysis, and prospective validation require a separate study. Finally, the data originate from Chinese clinical materials while the paper mainly presents English-rendered prompts; translation, demographic, and cultural bias may limit transfer. The release will include original-language prompts and evaluation scripts to support independent assessment.

\section{Ethical Considerations}
MIND is research-only screening support, not a diagnostic device or medical advice. LLM outputs can be unsupported; PRB supports are cues, not clinical evidence, and clinicians retain responsibility. Simulators and profiles are benchmarks, not patients. In five high-stakes scenarios---suicidal ideation, self-harm intent, psychosis crisis, medication requests, and emergency escalation---most sampled vignettes defer to clinicians or crisis services and avoid medication or dosage advice (Appendix~\ref{app:safety_evaluation}). Released artifacts are de-identified synthetic trajectories. Deployment requires prospective validation, human oversight, regulatory review, and bias and longitudinal-safety audits.

%\bibliography{anthology,mind_references}
% Deduplicated bibliography entries cited by this paper only
\bibliography{mind_references}

\appendix

\section{Turn-level Execution Procedure}
Algorithm~\ref{alg:mind_turn} gives the turn-level execution procedure used by MIND. At each turn, the dialogue history is first converted into a clinical retrieval state, whose observed evidence, missing criterion checks, and active differentials determine which PRB entries are admitted into the typed evidence state. The same state then conditions action generation, process-level validation, and controlled rectification, so retrieval is not an auxiliary context but the decision interface shared by policy and safeguards. This operationalizes the main paper's abstraction: long-horizon psychiatric inquiry is guided by criterion-linked evidence and reliability metadata rather than by unstructured retrieved text or prompt-only heuristics.

\begin{algorithm}[t]
\small
\DontPrintSemicolon
\caption{MIND turn execution with typed evidence-state construction and state-conditioned rectification}
\label{alg:mind_turn}
\KwInput{Dialogue history $H_t$, PRB, policy $\pi_\theta$, thresholds $\tau_{\rm rel}$, turn budget $L$}
\KwOutput{Doctor action $a_t$}
Generate clinical retrieval state $q_t$ from $H_t$\;
Parse $q_t$ into observed evidence $\mathcal{O}_t$, missing checks $\mathcal{M}_t$, and active differentials $\mathcal{D}_t$\;
Retrieve candidate PRB entries $\widetilde{\mathcal{S}}_t$ from PRB\;
Compute reliability $\mathrm{Rel}(q_t,e_i)$ for each entry\;
Set $\mathcal{S}_t=\{e_i:\mathrm{Rel}(q_t,e_i)\ge \tau_{\rm rel}\}$ and store scores/flags in $\mathcal{R}_t$\;
Construct $\mathcal{E}_t=(\mathcal{O}_t,\mathcal{M}_t,\mathcal{D}_t,\mathcal{S}_t,\mathcal{R}_t)$\;
Generate reasoning and action $a_t \sim \pi_\theta(\mathcal{E}_t,H_t)$\;
\If{$a_t$ violates format, repetition, safety, retrieval, or budget checks}{
    Retry once with the detected failure mode specified\;
    \If{retry still violates checks and fallback cap is not reached}{
        Replace $a_t$ with a structured reference inquiry from the nearest reliable PRB entry\;
    }
}
\Return{$a_t$}\;
\end{algorithm}

\begin{table}[t]
\centering
\caption{Support-faithfulness scores (0--10) on factual consistency (FC), support grounding (SG), and patient faithfulness (PF); DeepSeek-V3.2 judge. An independent GPT-4o cross-judge preserves the ranking (Table~\ref{tab:gpt4o_support_faithfulness}, Appendix~\ref{app:judge_robustness}).}
\label{tab:support_faithfulness_eval}
\vspace{-1mm}
\resizebox{0.48\textwidth}{!}{
\begin{tabular}{lcccc|lcccc}
\toprule
\textbf{Model} & \textbf{FC} & \textbf{SG} & \textbf{PF} & \textbf{Avg.} &
\textbf{Model} & \textbf{FC} & \textbf{SG} & \textbf{PF} & \textbf{Avg.} \\
\midrule
Qwen3-8B               & \large{6.0} & \large{5.6} & \large{5.9} & \large{5.8} &
Qwen3-8B$^{\dagger}$   & \large{7.0} & \large{7.1} & \large{6.8} & \large{7.0} \\
Baichuan-M2            & \large{6.1} & \large{5.7} & \large{5.9} & \large{5.9} &
Qwen3-32B$^{\dagger}$  & \large{7.3} & \large{7.4} & \large{7.1} & \large{7.3} \\
MRD-RAG                & \large{6.4} & \large{6.5} & \large{5.9} & \large{6.3} &
DoctorAgent-RL         & \large{7.4} & \large{7.5} & \large{7.0} & \large{7.3} \\
DDO                    & \large{7.5} & \large{7.7} & \large{7.3} & \large{7.5} &
Ours (MIND)            & \large{\textbf{8.3}} & \large{\textbf{8.5}} & \large{\textbf{8.0}} & \large{\textbf{8.3}} \\
\bottomrule
\end{tabular}
}
\vspace{-2mm}
\end{table}

\section{The Details of Experimental Setup}

\label{app_es}

\textbf{Datasets.}

Our dataset starts from 1{,}600 de-identified EMRs containing structured psychiatric descriptions, including chief complaints, symptom clusters, course/onset, severity and functional impact, and risk/exclusion cues. Each case is assigned to one of four ICD-10-style psychiatric screening categories~\cite{world1992icd}: Depression, Anxiety, Mixed Anxiety--Depression, and Other. The Mixed category denotes cases where both depressive and anxiety symptoms are clinically salient and the primary syndrome is difficult to determine from available evidence, while Other includes remaining psychiatric conditions or cases that cannot be confidently assigned to the first three categories. The raw EMR label distribution is 320/536/424/320 for Anxiety/Depression/Mix/Other. EMRs are split at the EMR level into 1{,}200 training and 400 held-out test EMRs; the EMRs themselves provide the clinical basis and screening labels, while simulated dialogues instantiate controlled multi-turn interactions for policy learning and evaluation.

Following MDD-5k-style construction, we reconstruct a structured patient profile from each EMR and simulate multi-turn consultations conditioned on the profile. Importantly, we enforce EMR-level split isolation \emph{before} profile reconstruction, dialogue simulation, PRB construction, and model training: the 1{,}200 training EMRs are the sole source of SFT data, RL trajectories, and PRB entries, while the 400 test EMRs feed only the closed-loop evaluation. All artifacts derived from a test EMR---reconstructed profiles, simulated dialogues, retrieval states, support notes, and clinical rationales---are excluded from the PRB used during both training and inference, preventing test-derived information from leaking into retrieval, reward computation, or policy optimization.

The resulting augmented data are summarized in Table~\ref{tab:data_stats}. The three rows are reported under \emph{different counting units} that match how each split is consumed: SFT is counted at the \emph{dialogue-turn} level---every (history, doctor-response) pair from the simulated training-EMR consultations is a separate supervised example---while RL and Test are counted at the \emph{dialogue} level, with each entry corresponding to one closed-loop trajectory rolled out from a training or test EMR under PsySim-Std/PsySim-Adapt with the standard turn budget $L\!=\!10$. The SFT row is therefore numerically larger than the underlying 1{,}200 training EMRs (an average of \(\sim\!11.6\) doctor turns per simulated consultation), the RL row reflects multi-seed roll-outs over the same 1{,}200 training EMRs across both simulators, and the Test row is the 602 dialogue-level evaluation instances obtained from the 400 held-out EMRs. The SFT average of $\sim\!11.6$ doctor turns is not constrained by the evaluation turn budget $L\!=\!10$, which applies only to closed-loop RL and test-time rollouts. Since high-quality psychiatric consultation data are scarce and costly to construct, we use this four-way label space as a controlled testbed for studying multi-turn psychiatric screening and differential inquiry, rather than as a replacement for fine-grained clinical diagnosis. The released resources follow the language of the available clinical materials and expert evaluations.

\begin{table}[t]
\centering
\caption{Dataset split and augmentation statistics by diagnostic category. SFT counts are at the \emph{dialogue-turn} level (each doctor-response turn from a simulated training-EMR consultation is one example); RL counts are at the \emph{dialogue} level, obtained as multi-seed PsySim-Std/PsySim-Adapt roll-outs over the same 1{,}200 training EMRs; Test counts are dialogue-level evaluation instances generated from the 400 held-out EMRs.}
\vspace{-1mm}
\resizebox{0.45\textwidth}{!}{
\begin{tabular}{l|c|c|c|c|c}
\toprule
\multirow{2}{*}{\shortstack{Split}} & \multicolumn{4}{c|}{Category} & \multirow{2}{*}{\shortstack{Total}} \\
\cmidrule{2-5}
 & Anxiety & Depression & Mix & Other &  \\
\midrule
SFT  & \large{2{,}410} & \large{4{,}120} & \large{3{,}640} & \large{3{,}720} & \large{\textbf{13{,}890}} \\
RL   & \large{884} & \large{1{,}929} & \large{1{,}141} & \large{1{,}485} & \large{\textbf{5{,}439}} \\
Test & \large{115} & \large{182} & \large{140} & \large{165} & \large{\textbf{602}} \\
\bottomrule
\end{tabular}
}
\label{tab:data_stats}
\end{table}

\section{Planned Release of De-identified Consultation Trajectories}
\label{app:dataset_release}
We will release the Chinese data, source code, trained models, evaluation scripts, construction protocols including PRB construction, and expert-reviewed, de-identified synthetic consultation trajectories generated by MIND to facilitate reproducible research on criteria-grounded psychiatric consultation. Each released instance contains the reconstructed patient profile, multi-turn inquiry trajectory, final ICD-10-style category, retrieved PRB supports, and basic metadata such as turn length and evidence tags. The release is intended as a research resource for studying inquiry behavior and diagnosis in controlled criteria-grounded settings, rather than as a substitute for real clinical data.

\paragraph{Quality Control.}
To improve the reliability of the released trajectories, domain experts reviewed the construction protocol and audited representative samples for profile consistency, diagnostic-label consistency, criteria coverage, and inappropriate clinical claims. During generation, each trajectory is constrained by the reconstructed patient profile and the ICD-10-style category, and expert feedback is used to refine prompts and filtering rules for inconsistent or unsupported dialogues. To prevent leakage, all splits are performed at the original EMR level before profile reconstruction, dialogue simulation, PRB construction, SFT/RL training, and evaluation. Consequently, any artifact derived from a held-out EMR, including patient profiles, simulated dialogues, retrieval states, PRB support notes, and rationales, is excluded from the training split and from the PRB used during inference.

\section{Baselines and Implementation Details}
\label{app:baselines_impl}

\textbf{Baselines.}

We organize baselines into four categories.

\textit{General-purpose LLMs.}
GPT-4o~\cite{hurst2024gpt} and DeepSeek-V3.2~\cite{liu2024deepseek} are evaluated in an inference-only multi-turn consultation setting, serving as strong references for general instruction following and reasoning without task-specific training.

\textit{Medical LLMs.}
We include Qwen3-4B~\cite{yang2025qwen3}, Qwen3-8B, Qwen3-32B, and Baichuan-M2~\cite{dou2025baichuan} as representative open medical-capable LLMs. We include Baichuan-M2 because it is a publicly available medical-capable model with accessible checkpoints and reproducible inference settings at the time of our experiments. More recent or unavailable variants are not included when their checkpoints, inference interface, or licensing conditions prevent controlled reproduction. These models are evaluated directly under the same psychiatric consultation protocol to measure medical-domain capability without additional tuning.

\textit{Retrieval-augmented consultation baselines.}
DDT~\cite{yin2025mdd} is a guideline or criteria-tree grounded dialogue baseline that asks questions following structured diagnostic checkpoints. MRD-RAG~\cite{sun2025multi} augments multi-turn consultation with retrieved supports to reduce unsupported clinical claims and improve evidence-conditioned decisions.

\textit{Fine-tuned and RL-based consultation models.}
Qwen3-4B$^{\dagger}$ and Qwen3-8B$^{\dagger}$ are instruction-tuned with the same warm-start data as our method for a controlled comparison. HuatuoGPT-o1-8B is included as a medical reasoning baseline. DoctorAgent-RL~\cite{feng2026doctoragent} and DDO~\cite{jia2025ddo} optimize multi-turn inquiry policies with reinforcement-learning-style objectives to improve question selection and final diagnosis. Finally, we report ours, MIND-4B and MIND-8B.

``$\dagger$'' denotes using the same instruction-tuning data as our supervised warm-start for a controlled comparison. All methods are evaluated under the same multi-turn protocol, and results are reported under the most appropriate setting for each baseline.

\textbf{Experimental Setup.}

Our framework follows a multi-turn doctor--patient dialogue paradigm. The doctor agent is built on Qwen3 backbones and warm-started with supervised fine-tuning on consultation instruction data, yielding MIND-4B and MIND-8B. The warm-start data are derived from the same PRB construction pipeline, which produces retrieval states, typed support entries, and reference inquiry rationales from training-split cases. We then optimize the doctor agent with GRPO on 8 NVIDIA A100 GPUs.

During each consultation turn, MIND summarizes the dialogue into a clinical retrieval state, retrieves criterion-relevant PRB entries, assigns them to the typed fields of $\mathcal{E}_t$, and admits them only when retrieval reliability passes the predefined gating threshold. The policy then generates explicit state-to-action reasoning and produces either the next inquiry or the final diagnosis. The maximum number of interaction turns is capped at $L=10$. We allow early diagnosis when a model explicitly produces a valid final label. However, to ensure fair comparison across models with different stopping behaviors, we use the same maximum turn budget and evaluate the final valid prediction. Since early stopping is not the focus of this work and many baselines are not trained with calibrated stopping policies, we do not use average stopping turn as a primary metric.

The RL reward combines both rule-based and LLM-based signals. Rule-based signals include terminal diagnostic correctness, format validation, pollution sanitization, repeated-question penalties, retrieval gating, and information-gain rewards computed from newly extracted symptoms. LLM-based signals are used for higher-level rubric scoring of intermediate reasoning quality, including symptom analysis, differential considerations, and decision logic, as well as auxiliary interaction-quality checks. Thus, the reward is not a model-based judge reward alone: the largest-weight terminal diagnostic reward is computed from the ground-truth screening label, while the LLM judge contributes only rubric-based process and auxiliary quality signals. Retrieval reliability is used for support gating and reward shaping, while a state-conditioned rectification mechanism triggers self-retry or PRB-guided fallback when low-utility turns are detected.

Unless otherwise specified, MIND refers to the 8B variant in the following experiments. We evaluate MIND and all baselines under the same multi-turn protocol and report accuracy and macro-average F1 as primary metrics. Experiments are conducted under two patient simulators, PsySim-Std and PsySim-Adapt, to assess robustness under simulator variation. In addition, we conduct support-faithfulness evaluation using DeepSeek-V3.2 as an automatic judge along three dimensions: factual consistency, support grounding of key clinical claims, and faithfulness to patient context. Human evaluation further assesses consultation quality under a standardized rubric. Key hyperparameters for SFT, RL training, and inference/simulation are summarized in Tables~\ref{tab:sft_params}, \ref{tab:rl_params}, and \ref{tab:inf_params}, respectively. The data distribution of the PRB support library is shown in Table~\ref{tab:support_quality_dist}.

\textbf{Statistical Testing.}
For accuracy and macro-F1, we estimate uncertainty with non-parametric bootstrap resampling over test cases. Specifically, we sample test cases with replacement for $B=1{,}000$ bootstrap replicates, recompute each metric on every replicate, and report the 2.5th and 97.5th percentiles as the 95\% confidence interval. To compare MIND with the strongest baseline under the same evaluation cases, we use paired bootstrap resampling: for each replicate, we resample cases once and compute the metric difference between the two systems on the identical resampled set. A one-sided $p$-value is estimated as the fraction of replicates where the difference is less than or equal to zero. We mark improvements as statistically significant when $p<0.05$.
In the current run, MIND improves over the strongest non-MIND baseline with statistically significant margins under both simulators. Under PsySim-Std, the accuracy gain is +8.8 points (95\% CI: [5.4, 12.1], $p<0.01$), and the macro-F1 gain is +8.7 points (95\% CI: [5.2, 12.0], $p<0.01$). Under PsySim-Adapt, the accuracy gain is +7.3 points (95\% CI: [3.9, 10.6], $p<0.01$), and the macro-F1 gain is +7.7 points (95\% CI: [4.1, 11.2], $p<0.01$).

\section{State-Representation Comparison}
\label{app:state_representation_comparison}

To isolate whether the gain comes from the evidence-state interface rather than from retrieval text or prompt formatting alone, we compare six policy/reward state-sharing variants under the same Qwen3-8B backbone, SFT warm-start data, GRPO configuration, reward weights, patient simulators, and turn budget. All six control rows disable state-conditioned fallback so that the comparison focuses on the representation supplied to the policy. Two of these controls (Policy=ES, Reward=Raw and Policy=Raw, Reward=ES) share the typed state with only the policy or only the reward to isolate policy--reward consistency. All variants use the same training cases, rollout budget, terminal reward, process-reward prompt, and evaluation scripts. Thus, this experiment changes the object of policy and reward conditioning while holding the optimization and interaction environment fixed.

\textit{Raw-history RL} conditions directly on the dialogue history $H_t$ and does not use PRB retrieval at inference time. \textit{Standard RAG RL} conditions on $H_t$ plus top-$k$ retrieved PRB support notes, but it appends the notes as prompt context and does not expose explicit observed-evidence, missing-check, differential-hypothesis, or reliability fields to the policy. \textit{Structured-prompt RAG RL} uses the same typed layout as $\mathcal{E}_t$---organizing retrieved supports into labeled observed/missing/differential/support/reliability sections in the policy prompt---but restricts this structure to the policy input alone; the process-reward judge and fallback operator receive raw history rather than the typed state. This variant isolates whether the gain comes from prompt formatting or from sharing the typed state across decision operators. \textit{Evidence-state RL} conditions on $\mathcal{E}_t=(\mathcal{O}_t,\mathcal{M}_t,\mathcal{D}_t,\mathcal{S}_t,\mathcal{R}_t)$: $\mathcal{O}_t$ contains patient-reported facts, $\mathcal{M}_t$ contains missing or unclear criterion checks, $\mathcal{D}_t$ contains active differential hypotheses, $\mathcal{S}_t$ contains reliability-gated support entries, and $\mathcal{R}_t$ contains support scores and hard-issue flags. The same typed state is shared with the process-reward judge and information-gain calculator. The final row reports full MIND with the same evidence-state interface plus state-conditioned rectification.

\begin{table*}[t]
\centering
\caption{Matched-variable controls over policy/reward state sharing. Acc/F1 are reported under the two patient simulators; MCR and repetition are averaged over inquiry turns. The six control rows isolate the representation by using the same backbone, rewards, and turn budget without state-conditioned fallback. Two one-sided rows (Policy=ES, Reward=Raw and Policy=Raw, Reward=ES) share the typed state with only policy or only reward to isolate policy--reward consistency. ``Structured-prompt RAG RL'' uses the same typed layout as $\mathcal{E}_t$ in the policy prompt but does not share it with the reward judge or fallback operator.}
\label{tab:state_representation_comparison}
\vspace{-1mm}
\resizebox{0.95\textwidth}{!}{
\begin{tabular}{l|cc|cc|cc}
\toprule
\multirow{2}{*}{\textbf{Variant}} &
\multicolumn{2}{c|}{\textbf{PsySim-Std}} &
\multicolumn{2}{c|}{\textbf{PsySim-Adapt}} &
\multicolumn{2}{c}{\textbf{Inquiry Quality}} \\
\cmidrule(lr){2-3}\cmidrule(lr){4-5}\cmidrule(lr){6-7}
& \textbf{Acc} & \textbf{Macro-F1} & \textbf{Acc} & \textbf{Macro-F1} & \textbf{MCR $\uparrow$} & \textbf{Rep. $\downarrow$} \\
\midrule
Raw-history RL & 63.4 & 64.5 & 53.9 & 54.4 & 43.5 & 14.5 \\
Standard RAG RL & 66.0 & 65.8 & 56.5 & 56.9 & 48.3 & 12.2 \\
Structured-prompt RAG RL & 66.7 & 66.5 & 57.2 & 57.5 & 49.5 & 11.5 \\
Policy=ES, Reward=Raw    & 67.2 & 67.2 & 57.8 & 58.0 & 50.8 & 10.2 \\
Policy=Raw, Reward=ES    & 66.4 & 66.0 & 56.9 & 57.0 & 49.0 & 11.8 \\
Evidence-state RL & 68.2 & 67.5 & 59.0 & 58.5 & 53.0 & 8.8 \\
\midrule
MIND & \textbf{69.0} & \textbf{68.8} & \textbf{59.8} & \textbf{59.3} & \textbf{54.2} & \textbf{8.0} \\
\bottomrule
\end{tabular}
}
\vspace{-2mm}
\end{table*}

The comparison shows a generally increasing trend as the policy input becomes more structured, but the gains are not uniform across the structuring steps. Raw-history RL performs reasonably but must repeatedly infer known facts, missing criteria, active alternatives, and reliability from a growing dialogue context. Standard RAG RL improves both diagnostic and inquiry metrics by exposing the policy to relevant supports, but it still treats retrieved text as unstructured context and cannot explicitly suppress weakly matched evidence or bind it to a missing criterion. Structured-prompt RAG RL yields a modest further gain (+0.7 Acc, +0.7 F1 under PsySim-Std) by organizing the prompt into typed sections, confirming that structured formatting helps the policy parse retrieved information. However, the gap from Structured-prompt RAG RL to Evidence-state RL (+1.5 Acc, +1.0 F1 under PsySim-Std; +1.8 Acc, +1.0 F1 under PsySim-Adapt) is substantially larger than the formatting-only gain, indicating that the main benefit comes from sharing the typed state with the reward judge and information-gain calculator, not from prompt layout alone. The full MIND row adds state-conditioned rectification on top of the evidence-state interface, yielding the best repetition rate and the strongest overall performance. These results support the central claim that the main novelty is not retrieval content or prompt formatting, but retrieval-as-state-construction with shared operational semantics across policy, reward, and recovery.

\section{Automatic Missing-Check Resolution Analysis}
\label{app:mcr_details}

We compute MCR using the structured retrieval state produced at each turn, without introducing additional human labels. Each retrieval state follows the fields used by the PRB construction prompt, including chief complaint, symptoms and duration, severity, sleep, impairment, safety risk, psychosis/mania cues, stressors, and substance use. A field is treated as a missing check when it is explicitly marked as ``not mentioned'', ``unclear'', or an equivalent null-status phrase in $q_t$. After the next patient response, we reconstruct $q_{t+1}$ and count the turn as resolving a missing check if at least one previously missing field becomes observed, explicitly negated, or otherwise clinically specified. Repetition is computed by semantic duplicate detection between the current inquiry and previous doctor questions in the same trajectory.

Formally, let $\mathcal{M}_t$ be the set of missing checks extracted from $q_t$, let $\Delta_t$ be the subset of those checks whose status changes after the next patient response, and let $T$ be the number of inquiry turns. We compute MCR only over turns that contain at least one missing check:
\begin{equation}
  \scriptstyle
\mathrm{MCR} =
\frac{
\sum_{t=1}^{T}
\mathbb{I}\!\left(|\mathcal{M}_t|>0\right)
\mathbb{I}\!\left(|\Delta_t|>0\right)
}{
\sum_{t=1}^{T}
\mathbb{I}\!\left(|\mathcal{M}_t|>0\right)
},
\end{equation}
where the denominator is the number of valid inquiry turns. This metric is intentionally defined over missing \emph{checks} rather than formal diagnostic criteria: the extracted gaps are operational fields in the evidence state, not manually annotated ICD-10 or DSM criterion labels. Thus, MCR measures whether the next question helps fill an explicit evidence gap recorded by the system.

\begin{table}[t]
\centering
\caption{Missing-check resolution under two rubrics. \emph{MCR-shared} applies MIND's missing-check schema as a unified rubric to every system; \emph{MCR-self} uses each system's own retrieval state (or extracted symptom mentions for baselines without one). Repetition is averaged over inquiry turns.}
\label{tab:mcr_analysis}
\vspace{-1mm}
\resizebox{0.48\textwidth}{!}{
\begin{tabular}{lccc}
\toprule
\textbf{Method} & \textbf{MCR-shared $\uparrow$} & \textbf{MCR-self $\uparrow$} & \textbf{Repetition $\downarrow$} \\
\midrule
Qwen3-8B               & 29.4 & 24.5 & 18.4 \\
MRD-RAG                & 32.8 & 35.4 & 15.6 \\
Qwen3-8B$^{\dagger}$   & 36.5 & 39.2 & 13.7 \\
DoctorAgent-RL         & 39.6 & 43.2 & 12.7 \\
DDO                    & 41.2 & 44.7 & 12.1 \\
MIND-8B                & \textbf{49.6} & \textbf{54.2} & \textbf{8.0} \\
\bottomrule
\end{tabular}
}
\vspace{-2mm}
\end{table}

\section{Judge Robustness for Support-Faithfulness Evaluation}
\label{app:judge_robustness}

The main support-faithfulness evaluation uses DeepSeek-V3.2 as the automatic judge. To reduce the risk that the conclusion is specific to the judge used for PRB construction or evaluation, we repeat the same evaluation with GPT-4o~\cite{hurst2024gpt}, which is from a different model family and is not used for PRB construction, SFT, RL rewards, model selection, or hyperparameter tuning. The prompt, sampled outputs, retrieved supports, patient context, and scoring dimensions are kept identical to Appendix Table~\ref{tab:support_faithfulness_eval}. GPT-4o assigns scores for factual consistency (FC), support grounding (SG), and patient faithfulness (PF) on the same 0--10 scale.

\begin{table}[t]
\centering
\caption{Support-faithfulness evaluation with GPT-4o as an independent judge from a different model family. DeepSeek-V3.2 is excluded from the evaluated systems because it serves as the primary judge in Table~\ref{tab:support_faithfulness_eval}. The ranking under GPT-4o is consistent with the DeepSeek-V3.2-judged results.}
\label{tab:gpt4o_support_faithfulness}
\vspace{-1mm}
\resizebox{0.48\textwidth}{!}{
\begin{tabular}{lcccc}
\toprule
\textbf{Model} & \textbf{FC} & \textbf{SG} & \textbf{PF} & \textbf{Avg.} \\
\midrule
Qwen3-8B               & 5.9 & 5.4 & 5.8 & 5.7 \\
Qwen3-8B$^{\dagger}$   & 6.9 & 7.0 & 6.7 & 6.9 \\
Qwen3-32B              & 6.0 & 5.7 & 5.9 & 5.9 \\
Qwen3-32B$^{\dagger}$  & 7.2 & 7.3 & 7.0 & 7.2 \\
Baichuan-M2            & 6.0 & 5.6 & 5.8 & 5.8 \\
MRD-RAG                & 6.3 & 6.4 & 5.8 & 6.2 \\
DoctorAgent-RL         & 7.3 & 7.4 & 6.9 & 7.2 \\
DDO                    & 7.4 & 7.6 & 7.2 & 7.4 \\
\textbf{MIND}          & \textbf{8.1} & \textbf{8.2} & \textbf{7.9} & \textbf{8.1} \\
\bottomrule
\end{tabular}
}
\vspace{-2mm}
\end{table}

\begin{table}[t]
\centering
\caption{Agreement between DeepSeek-V3.2 and GPT-4o support-faithfulness judgments. Pearson and Spearman correlations are computed at the sample level; weighted $\kappa$ is computed after rounding scores to integer bins.}
\label{tab:judge_agreement}
\vspace{-1mm}
\resizebox{0.42\textwidth}{!}{
\begin{tabular}{lccc}
\toprule
\textbf{Dimension} & \textbf{Pearson} & \textbf{Spearman} & \textbf{Weighted $\kappa$} \\
\midrule
FC   & 0.93 & 0.91 & 0.71 \\
SG   & 0.94 & 0.92 & 0.74 \\
PF   & 0.92 & 0.90 & 0.69 \\
Avg. & 0.95 & 0.93 & 0.76 \\
\bottomrule
\end{tabular}
}
\vspace{-2mm}
\end{table}

As shown in Tables~\ref{tab:gpt4o_support_faithfulness} and~\ref{tab:judge_agreement}, GPT-4o yields the same overall and per-dimension ordering as DeepSeek-V3.2, with MIND obtaining the highest average support-faithfulness score (8.1) and a $0.7$-point margin over the strongest baseline DDO; absolute scores from GPT-4o are uniformly $0.1$--$0.3$ points lower than those from DeepSeek-V3.2 across systems, so the rank-level gaps in Table~\ref{tab:support_faithfulness_eval} are reproduced rather than compressed. The high Pearson and Spearman correlations indicate that the relative differences among systems are stable across judge families, while the rounded-score weighted $\kappa$ values indicate substantial agreement. These results suggest that the support-faithfulness gains are not merely an artifact of a single evaluator's preferences.

\section{PRB Construction and Policy Prompts}
\label{app:prb_construction_prompts}

For reproducibility we list the offline prompts used to populate the typed PRB schema described in Section~\ref{sec:prb} from training-split dialogues, together with the Step~2 reasoning prompt used by the doctor policy at inference time.

\paragraph{Clinical Retrieval State $q_i$.}
\textit{\textbf{Prompt}: Compress the dialogue history into a fact-only clinical retrieval state $q_i$. Mark missing information explicitly and do not provide a diagnosis. Cover: chief complaint, symptoms and duration, severity, sleep, impairment, safety risk, psychosis/mania cues, stressors, and substance use. \textbf{Output}: One concise paragraph: ``Patient reports ...; duration ...; impairment ...; risk ...; psychosis/mania cues ...; stressors ...; substances ...''.}

\paragraph{Clinical Support Synthesis.}
\textit{\textbf{Prompt}: Given the dialogue history and clinical references, construct a criterion-indexed evidence entry. \textbf{Output}: (Criterion/check) ...; (Observed evidence) ...; (Missing or unclear check) ...; (Exclusion/safety cue) ...; (Next-inquiry rationale) Asking ``{next\_question}'' helps clarify ... and guides the next diagnostic decision.}

\paragraph{Reliability Assessment.}
\textit{\textbf{Prompt}: Verify whether the support is consistent with the dialogue context and clinically relevant to the references. Score reliability from 1 to 5 and mark hard issues. \textbf{Output}: (Score) 4/5. (Hard issues) H1--H3: NO. (Rationale) ...}

\paragraph{Doctor Policy Step~2 (state-to-action reasoning).}
\textit{\textbf{Prompt}: Given the dialogue history and the typed evidence state (observed evidence, missing checks, active differentials, reliable support entries, and reliability flags), reason about the current diagnostic state and decide the next action. \textbf{Output}: \texttt{<think>} The patient reports ..., while ... remains unclear. The key differential issue is .... The next question should clarify ... because it helps distinguish ... from .... \texttt{</think>} \texttt{<answer>} ... \texttt{</answer>}.}

\section{Sensitivity to PRB Construction LLMs}
\label{app:prb_construction_llm_sensitivity}

Although the main method only assumes an offline PRB construction procedure, we evaluate whether the choice of the off-the-shelf LLM used in this preprocessing step affects downstream performance. We instantiate the same PRB construction pipeline with Kimi-K2~\cite{team2025kimi}, DeepSeek-V3.2~\cite{liu2024deepseek}, and GLM-4.7~\cite{zeng2025glm}. All settings keep the same Qwen3-8B doctor backbone, PRB construction protocol, SFT data size, GRPO configuration, and inference budget. The only changed component is the preprocessing LLM used to construct retrieval states, support notes, and reference inquiry rationales from training-split cases.

\begin{table}[t]
\centering
\caption{Sensitivity to off-the-shelf LLMs used in the offline PRB construction pipeline. Scores are close across construction LLMs, with Kimi-K2 slightly higher in this run.}
\label{tab:prb_construction_llm_sensitivity}
\vspace{-1mm}
\resizebox{0.48\textwidth}{!}{
\begin{tabular}{lccccc}
\toprule
\multirow{2}{*}{\textbf{Construction LLM}} & \multirow{2}{*}{\textbf{Support}} & \multicolumn{2}{c}{\textbf{PsySim-Std}} & \multicolumn{2}{c}{\textbf{PsySim-Adapt}} \\
\cmidrule(lr){3-4}\cmidrule(lr){5-6}
& & \textbf{Acc} & \textbf{F1} & \textbf{Acc} & \textbf{F1} \\
\midrule
DeepSeek-V3.2~\cite{liu2024deepseek} & 8.22 & 68.3 & 68.0 & 59.1 & 58.5 \\
GLM-4.7~\cite{zeng2025glm} & 8.19 & 68.0 & 67.6 & 58.8 & 58.2 \\
Kimi-K2~\cite{team2025kimi} & \textbf{8.28} & \textbf{69.0} & \textbf{68.8} & \textbf{59.8} & \textbf{59.3} \\
\bottomrule
\end{tabular}
}
\vspace{-2mm}
\end{table}

As shown in Table~\ref{tab:prb_construction_llm_sensitivity}, performance remains stable across construction LLMs, indicating that the main gains come from criteria-grounded PRB usage, process supervision, state-conditioned rectification, and RL rather than a particular offline construction model. Kimi-K2 provides a small advantage, likely because its constructed supports are slightly more complete under our prompts, but the margin is modest.

\begin{table}[t]
\centering
\caption{Distribution of PRB support quality scores (1--5).}
\label{tab:support_quality_dist}
\vspace{-1mm}
\resizebox{0.45\textwidth}{!}{
\begin{tabular}{l c c c c c}
\toprule
\textbf{Score} & \textbf{1} & \textbf{2} & \textbf{3} & \textbf{4} & \textbf{5} \\
\midrule
\textbf{Count} & 2{,}318 & 27{,}597 & 15{,}459 & 47{,}628 & 8{,}676 \\
\textbf{Ratio} & 2.28\% & 27.14\% & 15.20\% & 46.84\% & 8.53\% \\
\bottomrule
\end{tabular}
}
\vspace{-2mm}
\end{table}

\begin{table}[t]
\centering
\begin{tabular}{l|c}
\hline
Hyperparameters & Value \\
\hline
Backbone model & Qwen3 \\
Training method & LoRA \\
LoRA rank ($r$) & 64 \\
LoRA alpha ($\alpha$) & 32 \\
Learning rate & 1e-4 \\
Warmup ratio & 0.03 \\
Effective batch size & 128 \\
Per-device batch size & 2 \\
Gradient accumulation steps & 8 \\
Training epochs & 3 \\
Max sequence length & 4096 \\
Gradient clipping & 1.0 \\
\hline
\end{tabular}
\caption{Key hyperparameters for the SFT warm-start stage. The effective batch size is obtained with gradient accumulation over 8 GPUs.}
\label{tab:sft_params}
\end{table}

\begin{table}[t]
\centering
\begin{tabular}{l|c}
\hline
Hyperparameters & Value \\
\hline
Algorithm & GRPO \\
Compute & 8$\times$ NVIDIA A100 \\
Actor learning rate & 5e-6 \\
KL coefficient ($\beta$) & 0.02 \\
Clip ratio ($\epsilon$) & 0.10--0.18 \\
Entropy coefficient & 1e-4 \\
Rollout backend & vLLM \\
Rollout temperature & 1.0 \\
Top-$p$ & 1.0 \\
Generations per prompt & 4 \\
Policy updates & 1{,}000--2{,}000 \\
Episodes per update & 128 \\
Mini-batch size & 64 \\
Micro-batch size per GPU & 1--2 \\
PPO epochs per update & 1 \\
Gradient clipping & 1.0 \\
Max prompt length & 4096 \\
Max response length & 1024 \\
\hline
Reward weights &  \\
Terminal screening reward & 5.0 \\
Information gain & 0.005 \\
Clinical reasoning & 0.01 \\
Format compliance & 0.1 \\
\hline
\end{tabular}
\caption{Key hyperparameters for RL policy optimization with GRPO. The update range reflects early stopping on validation performance and training stability.}
\label{tab:rl_params}
\end{table}

\begin{table}[t]
\centering
\caption{Operational triggers and caps for state-conditioned trajectory rectification. These checks are rule-based and are used both as intervention triggers and as penalty signals during RL.}
\label{tab:rectification_triggers}
\vspace{-1mm}
\resizebox{0.48\textwidth}{!}{
\begin{tabular}{p{0.28\textwidth}p{0.48\textwidth}p{0.18\textwidth}}
\toprule
\textbf{Trigger} & \textbf{Operational definition} & \textbf{Action} \\
\midrule
Malformed output & Missing required tags, invalid diagnosis format, or unparseable retrieval state. & Self-retry \\
Repeated inquiry & Semantic duplicate of a recent question or repeated check without new evidence. & Self-retry \\
Generic low-yield question & Question does not target symptoms, duration, impairment, safety, exclusion cues, or differential evidence. & Self-retry \\
Low retrieval reliability & No candidate support satisfies $\mathrm{Rel}(q_t,r_i)\ge\tau_{\mathrm{rel}}$. & Null support / fallback \\
Budget violation & Action ignores the remaining turn budget or fails to provide a final label at the last turn. & Penalty / forced diagnosis \\
Unsafe or off-track response & Response contains irrelevant, non-clinical, or unsafe patient-facing content. & Self-retry / fallback \\
\midrule
Self-retry cap & At most once per turn. & Skip to fallback \\
Fallback cap & At most twice per episode. & Force diagnosis next turn \\
Total intervention cap & At most three rectification interventions per episode. & Force diagnosis next turn \\
\bottomrule
\end{tabular}
}
\vspace{-2mm}
\end{table}

\begin{table*}[t]
\centering
\caption{Rubric for turn-level clinical process reward. Scores are assigned on a 0--2 scale for each dimension and normalized before reward aggregation.}
\label{tab:process_reward_rubric}
\vspace{-1mm}
\resizebox{\textwidth}{!}{
\begin{tabular}{p{0.16\textwidth}p{0.25\textwidth}p{0.25\textwidth}p{0.25\textwidth}}
\toprule
\textbf{Dimension} & \textbf{0} & \textbf{1} & \textbf{2} \\
\midrule
Symptom analysis & Missing or contradicting key symptoms & Mentions some relevant symptoms but omits important fields & Correctly identifies confirmed, negated, and missing symptom evidence \\
Differential focus & No clear differential or irrelevant alternatives & Partially considers competing categories & Explicitly distinguishes plausible categories and relevant exclusions \\
Decision logic & Next action is unsupported or generic & Action is somewhat relevant but weakly justified & Action directly targets the most informative missing evidence \\
\bottomrule
\end{tabular}
}
\vspace{-2mm}
\end{table*}

\begin{table}[t]
\centering
\begin{tabular}{l|c}
\hline
Hyperparameters & Value \\
\hline
Doctor agent (MIND) &  \\
Sampling temperature & 1.0 \\
Top-$p$ & 1.0 \\
Max response length & 2048 \\
Reliability threshold $\tau_{\mathrm{rel}}$ & 0.2 \\
\hline
Patient simulator &  \\
Backbone model & Qwen3 (fixed) \\
Temperature & 0.0 \\
Max interaction turns ($L$) & 10 \\
Max response length & 512 \\
\hline
\end{tabular}
\caption{Hyperparameters for inference and patient-agent simulation.}
\label{tab:inf_params}
\end{table}

\begin{figure}[t]
\centering
\includegraphics[width=1\columnwidth, height=!]{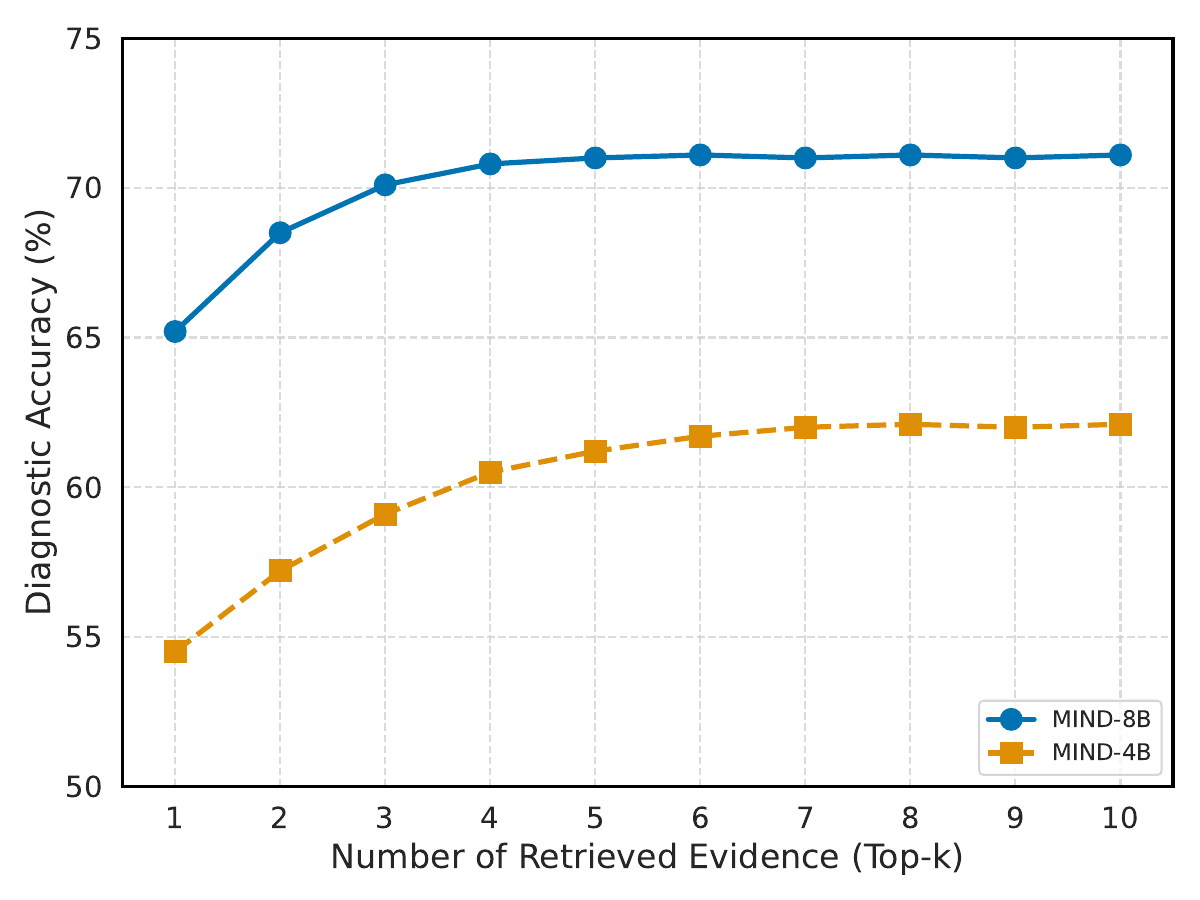}
\caption{Retrieval-depth sensitivity. Accuracy of MIND-4B/8B vs.\ Top-$k$.}
\label{para1}
\end{figure}

\begin{figure}[t]
\centering
\includegraphics[width=1\columnwidth, height=!]{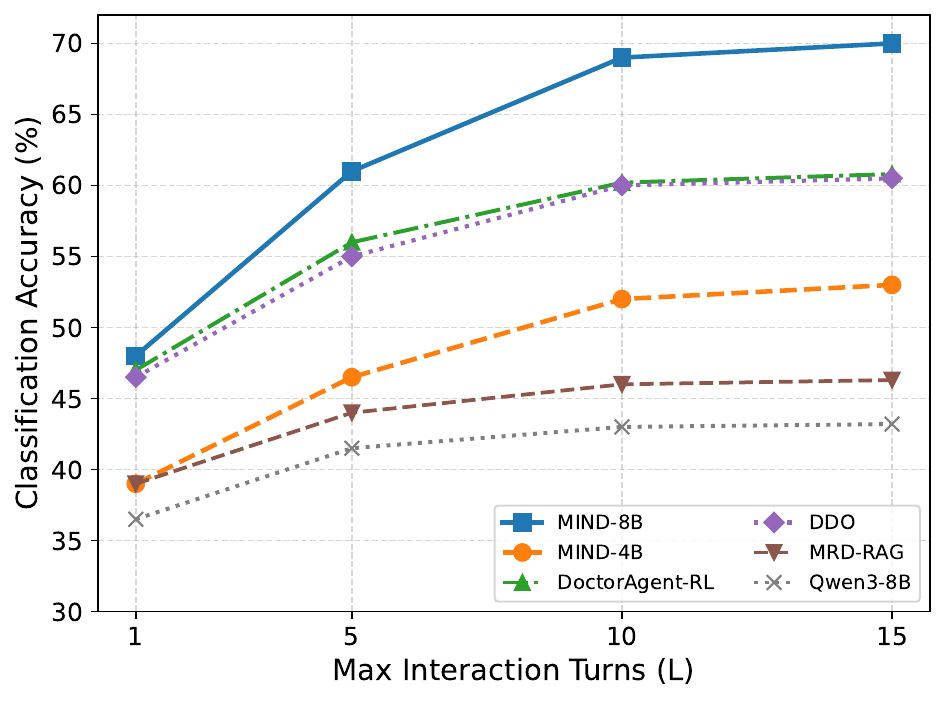}
\caption{Dialogue-horizon sensitivity. Accuracy of MIND and baselines vs.\ max turns $L$.}
\label{para2}
\end{figure}

\section{External Evaluation on MDD-5k}
\label{app:external_mdd5k}

\subsection{Diagnostic Evaluation on Public MDD-5k Dialogues}
\label{app:external_mdd5k_eval}

To further reduce dependence on our closed-loop simulator, we conduct an external dialogue-level diagnostic evaluation on the publicly released dialogue portion of MDD-5k. Although the original EMRs used to construct MDD-5k are not public, its released diagnostic dialogues provide a suitable public benchmark for testing whether a model can infer the final psychiatric screening category from dialogue evidence.

For each held-out dialogue, unless otherwise specified, the model receives the complete released dialogue history and predicts one of the four screening categories: Depression, Anxiety, Mixed Anxiety--Depression, or Other. The prediction is compared with the released diagnostic label provided by MDD-5k. All reported scores are produced by our re-evaluation under a unified prompt, label mapping, and evaluation protocol, rather than copied from the original MDD-5k report: released depressive diagnoses are mapped to Depression, anxiety diagnoses to Anxiety, anxiety--depression or co-occurring depressive/anxiety diagnoses to Mixed Anxiety--Depression, and all remaining diagnoses to Other. After mapping, the evaluation set contains $N{=}1{,}200$ dialogues with a four-way distribution of 380/340/260/220 (Depression/Anxiety/Mix/Other); 15 dialogues with unmappable labels are excluded. We report classification accuracy and macro-F1 on the mapped set. This evaluation set is used only for external testing and is excluded from PRB construction, SFT, RL training, reward design, simulator tuning, and hyperparameter selection.

\begin{table}[t]
\centering
\small
\caption{External held-out MDD-5k diagnostic evaluation. Models receive the full released dialogue and predict the four-way category; the set is excluded from PRB, SFT, RL, reward, simulator, and hyperparameter tuning.}
\label{tab:external_mdd5k_diag}
\vspace{-1mm}
\resizebox{0.48\columnwidth}{!}{
\begin{tabular}{lcc}
\toprule
Model & Accuracy & Macro-F1 \\
\midrule
Qwen3-8B & 34.8 & 32.9 \\
Qwen3-8B$^{\dagger}$ & 42.5 & 40.6 \\
MRD-RAG & 38.6 & 36.8 \\
DoctorAgent-RL & 46.8 & 45.0 \\
DDO & 48.6 & 47.0 \\
MIND-8B & \textbf{53.8} & \textbf{52.0} \\
\bottomrule
\end{tabular}
}
\vspace{-2mm}
\end{table}

The absolute scores are lower than the EMR-grounded simulator results in Table~\ref{tab:perclass_prf_patientsim}, which is expected because the models cannot ask additional follow-up questions and must diagnose from fixed evidence. The ranking is nevertheless consistent with the main results: retrieval-augmented, tuned, and RL-based baselines improve over inference-only generation, while MIND obtains the strongest external dialogue-level performance.

\subsection{Next-Question Evaluation on Public MDD-5k Dialogue Prefixes}
\label{app:mdd5k_prefix_next_question}

We further evaluate whether MIND can select clinically useful follow-up questions from public real-dialogue evidence. We sample 80 dialogue prefixes from the publicly released MDD-5k diagnostic dialogues, stratified across the four mapped categories (20 per category). For each case, the model is given the chief complaint or the first $K{=}10$ doctor--patient exchanges and is asked to generate one next doctor question. No simulated patient response is provided, and no further rollout is performed.
This setting isolates the immediate inquiry decision from the downstream simulator response. It therefore tests whether a model can identify a useful next evidence gap from public partial consultation context, rather than whether it can exploit simulator-specific response patterns. The same prefix is given to all systems, and PRB-based methods use only train-split PRB entries with reliability-gated support injection.

The generated questions are evaluated by three reviewers with medical or clinical training (one licensed psychiatrist and two medically trained graduate students) under an anonymized, randomized protocol. Each reviewer independently scores all model outputs for every prefix; model identities are hidden. We use a pre-defined rubric tailored to the four-category psychiatric screening setting. The rubric covers four aspects: clinical relevance, expected information gain, safety and exclusion awareness, and patient-facing naturalness. Clinical relevance measures whether the question targets symptoms or contextual factors that are meaningful for the observed prefix. Expected information gain measures whether the answer would likely reduce uncertainty among Depression, Anxiety, Mixed Anxiety--Depression, and Other. Safety and exclusion awareness measures whether the question appropriately checks risk, functional impairment, manic or psychotic cues, substance use, or other exclusion factors when such checks are indicated. Naturalness measures whether the question is understandable, non-leading, and acceptable in a patient-facing consultation.

Scores are assigned on a 1--5 scale for each dimension and averaged across reviewers. When reviewers disagree substantially, the case is discussed and, if necessary, adjudicated by an additional senior reviewer. This evaluation measures the quality of the immediate inquiry decision rather than final diagnostic accuracy.

\begin{table}[t]
\centering
\caption{Rubric for next-question evaluation on public MDD-5k dialogue prefixes. Scores are assigned on a 1--5 scale, where higher scores indicate better inquiry quality.}
\label{tab:mdd5k_prefix_rubric}
\vspace{-1mm}
\resizebox{0.48\textwidth}{!}{
\begin{tabular}{p{0.24\textwidth}p{0.68\textwidth}}
\toprule
\textbf{Dimension} & \textbf{Scoring criterion} \\
\midrule
Clinical relevance & Whether the question targets symptoms, context, duration, impairment, or other evidence relevant to the observed prefix. \\
Expected information gain & Whether the answer would likely reduce uncertainty among Depression, Anxiety, Mixed Anxiety--Depression, and Other. \\
Safety/exclusion awareness & Whether the question checks safety risk, functional impairment, psychosis/mania cues, substance use, or other exclusion factors when indicated. \\
Naturalness & Whether the question is understandable, non-leading, concise, and acceptable in a patient-facing consultation. \\
\bottomrule
\end{tabular}
}
\vspace{-2mm}
\end{table}

\begin{table}[t]
\centering
\caption{Next-question evaluation on public MDD-5k dialogue prefixes. Each model receives the same public chief complaint or first $K{=}10$ doctor--patient exchanges and generates one follow-up question. Scores are averaged across reviewers on a 1--5 scale.}
\label{tab:mdd5k_prefix_next_question}
\vspace{-1mm}
\resizebox{0.48\textwidth}{!}{
\begin{tabular}{lcccc}
\toprule
Method & Rel. & Info. & Safety & Nat. \\
\midrule
Qwen3-8B & 3.15 & 3.05 & 2.85 & 3.45 \\
MRD-RAG & 3.62 & 3.58 & 3.45 & 3.50 \\
DoctorAgent-RL & 3.68 & 3.61 & 3.48 & 3.60 \\
DDO & 3.70 & 3.66 & 3.55 & 3.62 \\
MIND-8B & \textbf{4.18} & \textbf{4.10} & \textbf{4.02} & \textbf{3.93} \\
\bottomrule
\end{tabular}
}
\vspace{-2mm}
\end{table}

This evaluation complements the external diagnostic task in Appendix~\ref{app:external_mdd5k_eval}. The former tests label prediction from released dialogue evidence, whereas this experiment tests whether a model selects an appropriate next inquiry from a public partial consultation. Together, the two analyses provide additional evidence about model behavior outside simulator-based interaction.

\section{Evidence-State Analyses}
\label{app:evidence_state_analyses}

\subsection{Evidence-State Field Masking Analysis}
\label{app:field_masking}

To verify that each field of the evidence-state interface has a distinct causal role in the decision process, we conduct a field-masking experiment at inference time. Starting from the full MIND system, we independently mask (replace with an empty placeholder) each of the five fields---$\mathcal{O}_t$, $\mathcal{M}_t$, $\mathcal{D}_t$, $\mathcal{S}_t$, $\mathcal{R}_t$---while keeping all other components unchanged. The same test cases, turn budget, and evaluation protocol are used. This experiment goes beyond standard component ablation (e.g., ``w/o PRB'') by isolating the contribution of individual state fields rather than entire modules.

\begin{table}[t]
\centering
\caption{Evidence-state field masking at inference time. Each row masks one field of $\mathcal{E}_t$ while keeping all other components unchanged. Acc and F1 are averaged over PsySim-Std and PsySim-Adapt; the Full MIND averages are computed from the fixed main-table scores. MCR and Support are reported under PsySim-Std.}
\label{tab:field_masking}
\vspace{-1mm}
\resizebox{0.48\textwidth}{!}{
\begin{tabular}{lcccc}
\toprule
\textbf{Variant} & \textbf{Acc} & \textbf{Macro-F1} & \textbf{MCR $\uparrow$} & \textbf{Support} \\
\midrule
Full MIND & \textbf{64.4} & \textbf{64.1} & \textbf{54.2} & \textbf{8.3} \\
\midrule
mask $\mathcal{O}_t$ (observed evidence) & 60.8 & 62.2 & 49.4 & 8.2 \\
mask $\mathcal{M}_t$ (missing checks) & 60.1 & 60.9 & 41.1 & 8.1 \\
mask $\mathcal{D}_t$ (active differentials) & 61.5 & 63.3 & 51.5 & 8.0 \\
mask $\mathcal{S}_t$ (criterion supports) & 61.2 & 62.5 & 47.9 & 7.6 \\
mask $\mathcal{R}_t$ (reliability metadata) & 62.6 & 63.8 & 52.6 & 7.7 \\
\bottomrule
\end{tabular}
}
\vspace{-2mm}
\end{table}

Table~\ref{tab:field_masking} shows that masking any single field degrades performance, but the degradation patterns differ across fields, confirming that the fields serve distinct operational roles rather than being redundant descriptions of the same information.

\textbf{$\mathcal{M}_t$ (missing checks)} produces the largest MCR drop ($-$13.1 points) and the largest accuracy drop ($-$4.3 points), consistent with its designed role as the inquiry-target selector: without explicit missing-check information, the policy loses its primary signal for choosing which criterion to pursue next, leading to more generic and repeated questions.

\textbf{$\mathcal{O}_t$ (observed evidence)} produces the second-largest accuracy drop ($-$3.6 points), because the policy can no longer anchor its diagnostic hypotheses in confirmed patient reports and must re-derive observed facts from the raw dialogue history.

\textbf{$\mathcal{S}_t$ (criterion supports)} produces the largest Support faithfulness drop ($-$0.7 points), confirming that criterion-linked support entries are the primary driver of grounded clinical claims. Without them, the policy generates more unsupported assertions.

\textbf{$\mathcal{D}_t$ (active differentials)} has a moderate effect on accuracy but a smaller effect on MCR, suggesting that differential hypotheses primarily influence which diagnostic category the policy considers rather than which question it asks next.

\textbf{$\mathcal{R}_t$ (reliability metadata)} has the smallest individual effect, but its removal still degrades Support faithfulness by $-$0.6 points, confirming that reliability gating prevents weakly matched supports from influencing the policy even when the other fields are intact.

These results provide causal evidence that the evidence-state interface is not a monolithic prompt template but a decomposed decision structure in which each field controls a different aspect of the consultation process.

\subsection{Evidence-State Construction Quality}
\label{app:state_construction_quality}

We evaluate the quality of evidence states produced by the state-construction operator $f_\phi$ on a manually inspected subset of test trajectories.

\paragraph{Protocol.}
We randomly sample 100 turns from 20 test-set consultation trajectories (5 turns per trajectory on average). For each turn, two medically trained annotators independently inspect the constructed $\mathcal{E}_t$ and annotate whether each field is (a) \emph{correct}: the field accurately reflects the dialogue history up to turn $t$; (b) \emph{partially correct}: the field captures the main content but omits or slightly misrepresents some details; or (c) \emph{incorrect}: the field contains fabricated, contradictory, or substantially wrong information. We report the fraction of turns rated as correct or partially correct (``Acc$_{\ge\text{partial}}$'') and the fraction rated as fully correct (``Acc$_{\text{full}}$''). Inter-annotator agreement is measured by Cohen's $\kappa$.

\begin{table}[t]
\centering
\caption{Evidence-state construction quality on 100 manually inspected turns. Acc$_{\ge\text{partial}}$ counts correct or partially correct; Acc$_{\text{full}}$ counts fully correct only.}
\label{tab:state_construction_quality}
\vspace{-1mm}
\resizebox{0.48\textwidth}{!}{
\begin{tabular}{lccc}
\toprule
\textbf{Field} & \textbf{Acc$_{\ge\text{partial}}$} & \textbf{Acc$_{\text{full}}$} & \textbf{Cohen's $\kappa$} \\
\midrule
$\mathcal{O}_t$ (observed evidence) & 94.0 & 82.0 & 0.78 \\
$\mathcal{M}_t$ (missing checks) & 91.0 & 76.0 & 0.73 \\
$\mathcal{D}_t$ (active differentials) & 89.0 & 71.0 & 0.70 \\
$\mathcal{S}_t$ (criterion supports) & 92.0 & 78.0 & 0.75 \\
$\mathcal{R}_t$ (reliability metadata) & 93.0 & 80.0 & 0.77 \\
\midrule
Overall & 91.8 & 77.4 & 0.75 \\
\bottomrule
\end{tabular}
}
\vspace{-2mm}
\end{table}

Table~\ref{tab:state_construction_quality} shows that the state constructor achieves high agreement with expert annotations across all five fields. Observed evidence ($\mathcal{O}_t$) and reliability metadata ($\mathcal{R}_t$) are the most accurately constructed, likely because they involve relatively straightforward extraction (patient statements) and scoring (reliability rubric). Active differentials ($\mathcal{D}_t$) have the lowest full-accuracy rate, reflecting the inherent ambiguity of differential reasoning in early-stage psychiatric consultations where multiple hypotheses are plausible. Missing checks ($\mathcal{M}_t$) achieve 91.0\% partial-or-better accuracy, indicating that the state constructor reliably identifies evidence gaps even when the exact phrasing differs from expert expectations.

These results support the claim that $\mathcal{E}_t$ is not merely a naming convention but a measurably accurate intermediate representation. The field-masking analysis in Appendix~\ref{app:field_masking} further shows how construction errors can affect downstream behavior: missing-check failures are most harmful for both accuracy ($-4.3$) and inquiry quality (MCR $-13.1$), observed-evidence failures cause the second-largest accuracy drop ($-3.6$), and support/reliability failures mainly reduce support faithfulness ($-0.7$ and $-0.6$). Active differentials are the hardest field to construct fully, so ambiguous early turns remain the main sensitivity point.

\section{Additional Fine-Grained Analyses}
\label{app:fine_grained_analyses}

Beyond aggregate screening-category metrics, we provide additional analyses to examine whether MIND improves criteria-level inquiry behavior, retrieval robustness, and trajectory stability. These analyses are intended to clarify how the proposed components affect the consultation process rather than to claim clinical deployment readiness.

\begin{table}[t]
\centering
\caption{High-overlap screening categories. Values are per-class F1 from Table~\ref{tab:perclass_prf_patientsim}; $\Delta$ compares MIND with the strongest non-MIND baseline for the same simulator and class.}
\label{tab:high_overlap_classes}
\vspace{-1mm}
\resizebox{0.48\textwidth}{!}{
\begin{tabular}{llccc}
\toprule
\textbf{Simulator} & \textbf{Class} & \textbf{Best non-MIND} & \textbf{MIND} & $\Delta$ \\
\midrule
PsySim-Std & Mix & 55.2 & 63.5 & +8.3 \\
PsySim-Adapt & Anxiety & 53.2 & 52.0 & -1.2 \\
PsySim-Adapt & Mix & 46.0 & 46.5 & +0.5 \\
\bottomrule
\end{tabular}
}
\vspace{-2mm}
\end{table}

Table~\ref{tab:high_overlap_classes} isolates high-overlap Anxiety/Mix boundaries. MIND improves Mix under both simulators but is slightly below DoctorAgent-RL on Anxiety under PsySim-Adapt, indicating that subjective anxiety--depression overlap remains a challenging boundary case despite the aggregate gains.

\begin{figure}[t]
\centering
\includegraphics[width=0.98\columnwidth]{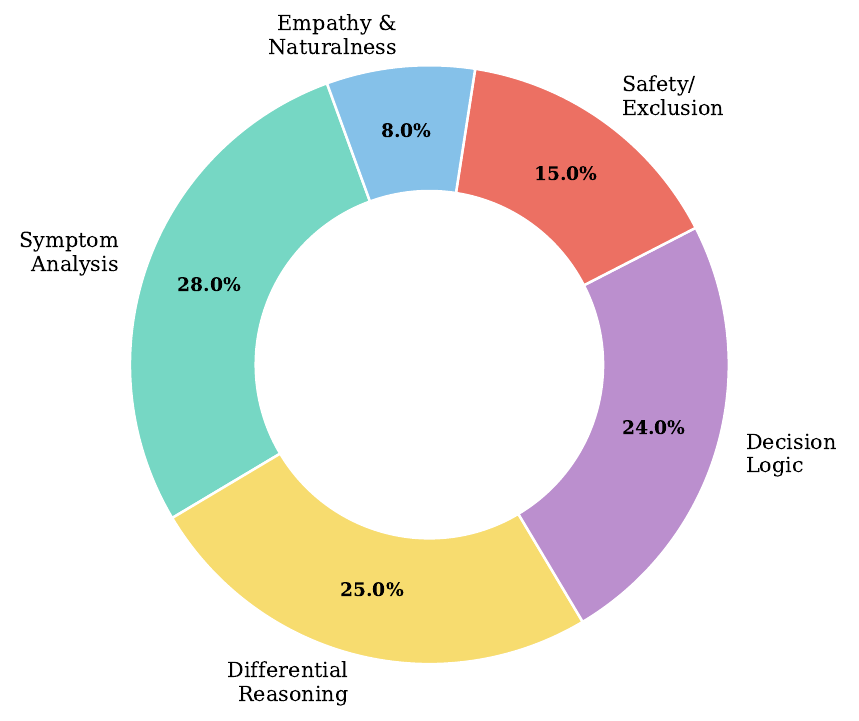}
\caption{Composition of MIND evaluation dimensions, centered on symptom analysis, differential reasoning, and decision logic, with auxiliary safety/exclusion and naturalness checks.}
\label{fig:evaluation_dimension_donut}
\end{figure}

\begin{figure}[t]
\centering
\includegraphics[width=0.98\columnwidth]{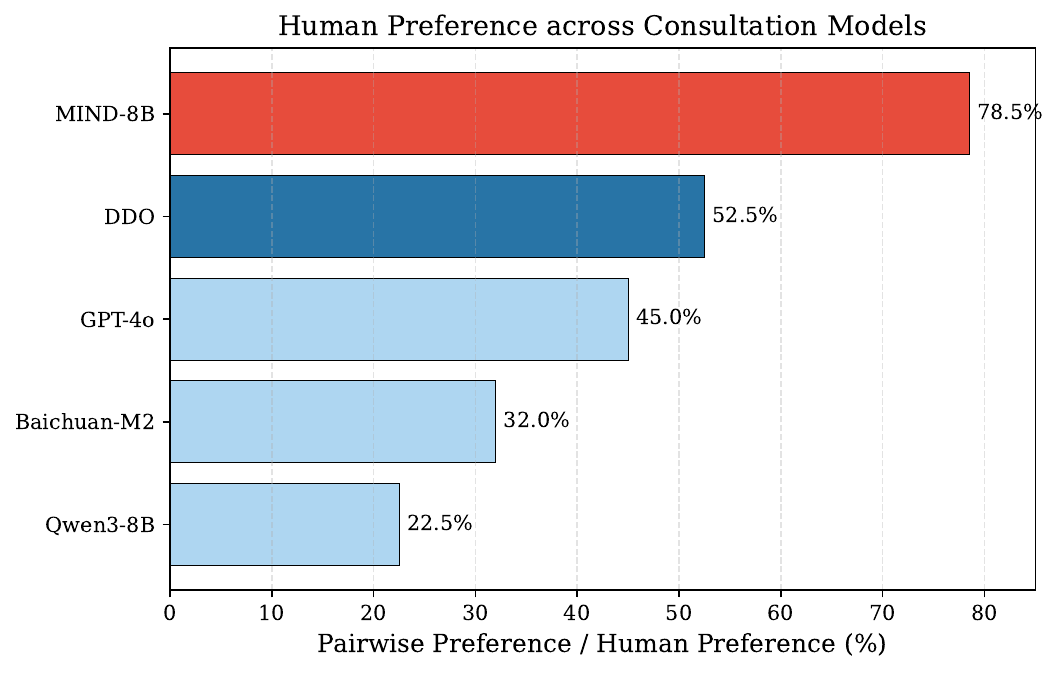}
\caption{Human preference across consultation models. MIND ranks first, suggesting criteria-grounded inquiry and rectification benefit patient-facing quality beyond label accuracy.}
\label{fig:horizontal_preference_accuracy}
\end{figure}

Figure~\ref{fig:evaluation_dimension_donut} summarizes the composition of the evaluation dimensions used in our fine-grained analyses. The largest portions correspond to symptom analysis, differential reasoning, and decision logic, reflecting the clinical process reward design, while safety/exclusion awareness and patient-facing naturalness provide auxiliary checks on interaction quality. Figure~\ref{fig:horizontal_preference_accuracy} reports a horizontal comparison of human preference scores across representative consultation models, highlighting whether improvements in automated metrics translate into preferred patient-facing behavior.

\begin{figure}[t]
\centering
\includegraphics[width=0.98\columnwidth]{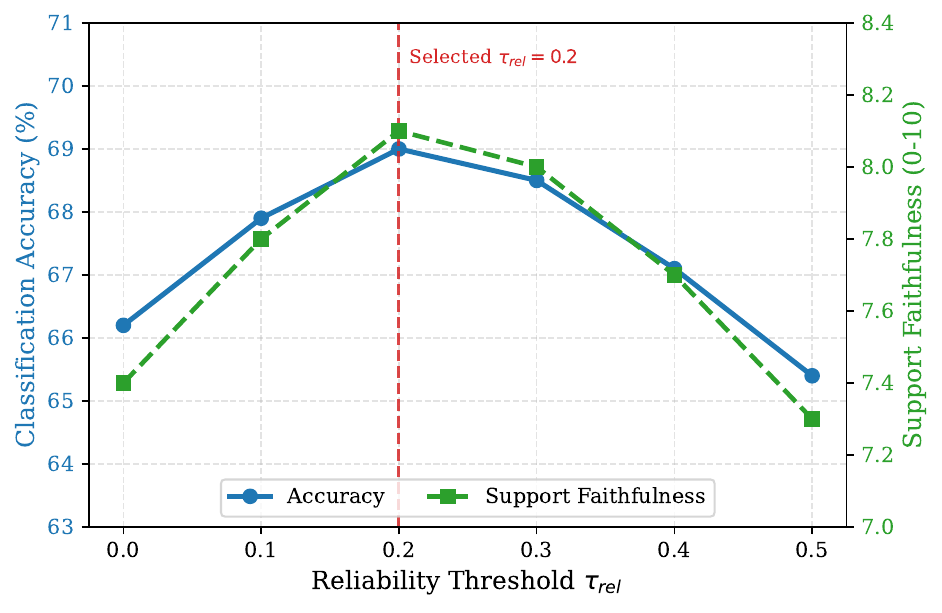}
\caption{Sensitivity to PRB reliability threshold $\tau_{\rm rel}$ (full version). Accuracy uses gold ICD-10-style labels; the faithfulness axis is scored by the \emph{independent GPT-4o cross-judge} (Table~\ref{tab:gpt4o_support_faithfulness}, Appendix~\ref{app:judge_robustness}), \emph{not} by the LLM that produced reliability scores $\rho_i$, so the inverted-U is not a self-judging artifact. Moderate gating ($\tau_{\rm rel}{=}0.2$) jointly maximizes both axes; overly permissive admits noisy supports, overly strict drops useful criteria guidance.}
\label{fig:rel_threshold_sensitivity}
\end{figure}

\begin{figure*}[t]
\centering
\includegraphics[width=0.92\textwidth]{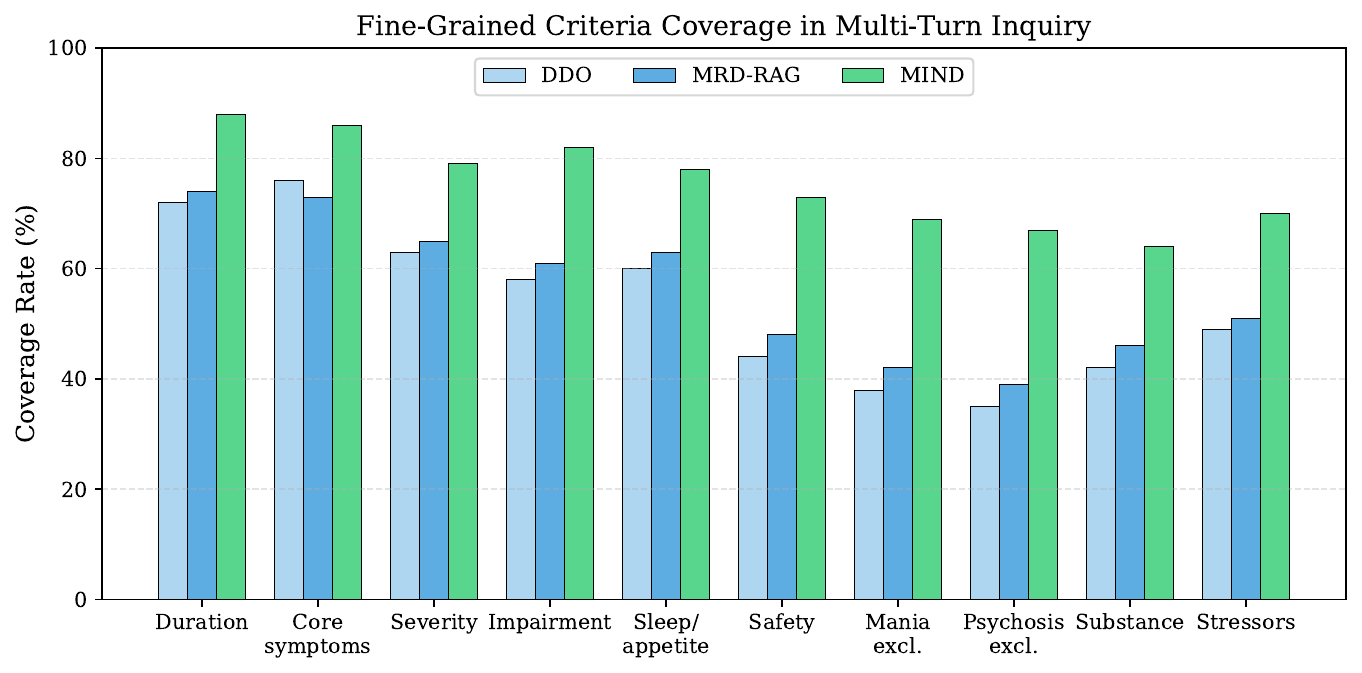}
\caption{Criteria coverage in multi-turn inquiry. MIND raises coverage of clinically decisive cues---functional impairment, safety risk, exclusion---suggesting more complete differential inquiry beyond label optimization.}
\label{fig:criteria_coverage}
\end{figure*}

\begin{figure}[t]
\centering
\includegraphics[width=0.98\columnwidth]{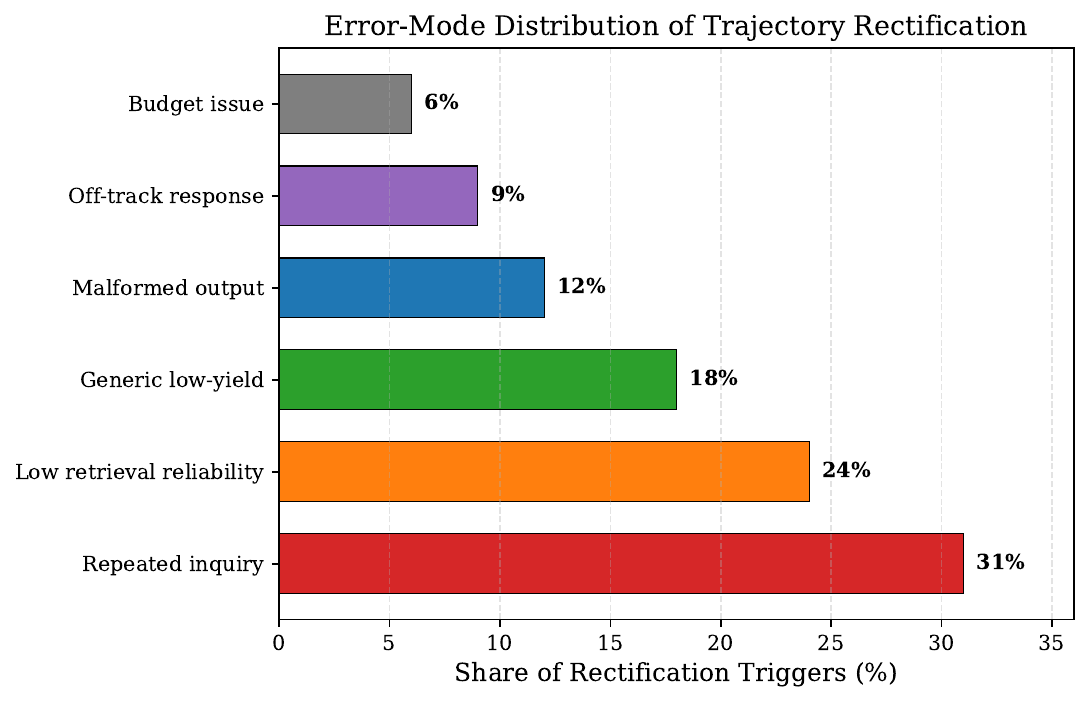}
\caption{Rectification trigger distribution. Most interventions stem from repetition, low reliability, or generic low-yield turns, showing rectification targets long-horizon drift rather than format errors alone.}
\label{fig:rectification_error_modes}
\end{figure}

\begin{figure}[t]
\centering
\includegraphics[width=0.98\columnwidth]{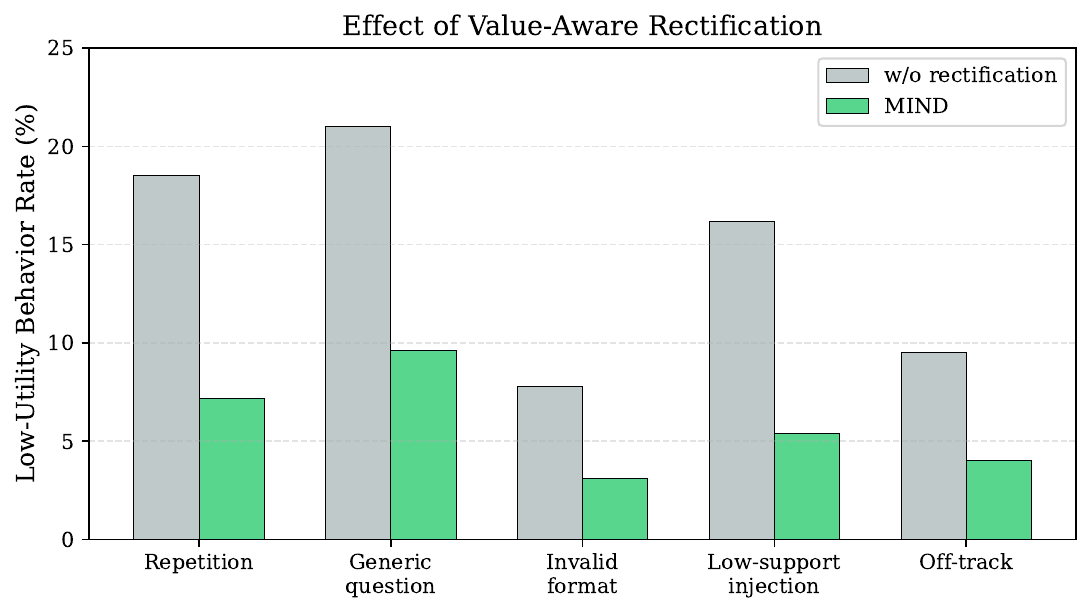}
\caption{Effect of state-conditioned rectification. MIND reduces repeated and generic questions, invalid outputs, and unsupported support injection vs.\ the no-rectification variant.}
\label{fig:before_after_rectification}
\end{figure}

\begin{figure}[t]
\centering
\includegraphics[width=0.98\columnwidth]{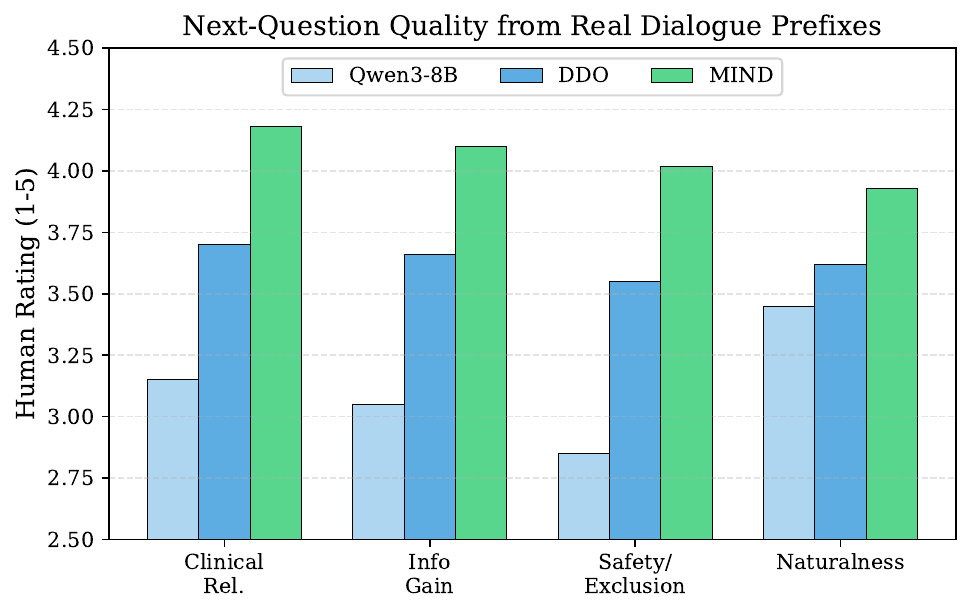}
\caption{Human-rated next-question quality on public MDD-5k prefixes. MIND gains most on information gain and safety/exclusion awareness, suggesting criteria-grounded inquiry transfers beyond simulator rollout.}
\label{fig:real_prefix_next_question_breakdown}
\end{figure}

\begin{figure}[t]
\centering
\includegraphics[width=0.98\columnwidth]{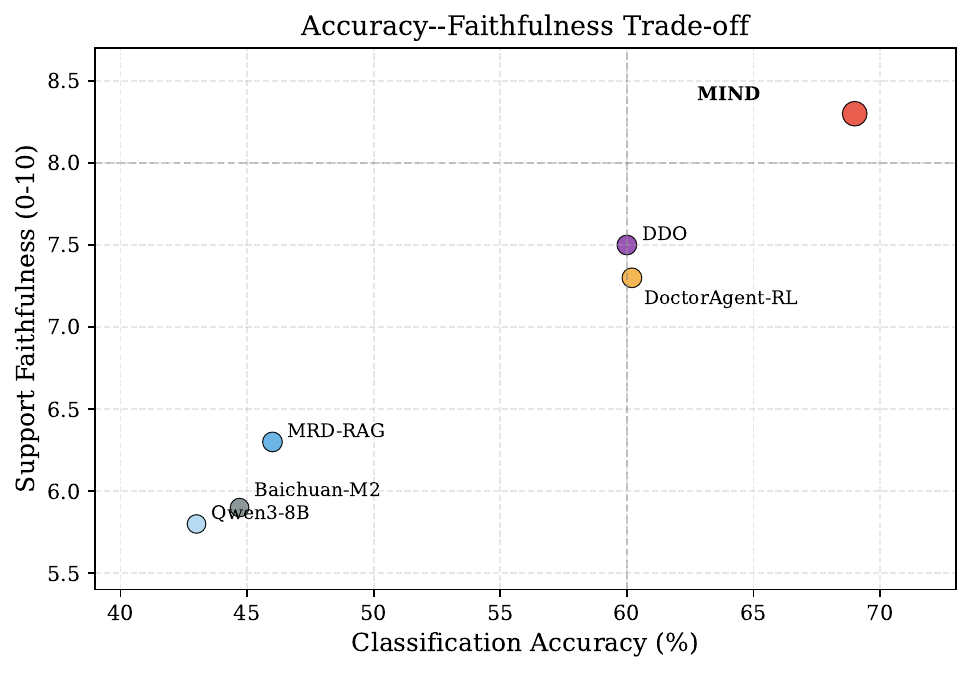}
\caption{Accuracy--faithfulness trade-off. MIND sits on the upper-right frontier: criteria-grounded retrieval improves screening while preserving stronger support faithfulness.}
\label{fig:accuracy_faithfulness_scatter}
\end{figure}

\section{Parameter and State-Variant Analyses}
\label{app:parameter_and_state_variants}

\subsection{Parameter Analysis}
Figures~\ref{para1}--\ref{para2} study the sensitivity of MIND to the retrieval depth and the dialogue horizon.
In Figure~\ref{para1}, increasing Top-$k$ consistently improves accuracy for both MIND-4B and MIND-8B, with the largest gains observed when moving from very shallow retrieval ($k{=}1$) to moderate retrieval ($k{\approx}3$--$5$). Beyond this range, the curves gradually saturate, suggesting diminishing returns once the core criteria-relevant supports have been covered; additional supports provide limited new discriminative signals and mainly serve as redundancy. In Figure~\ref{para2}, allowing more interaction turns also yields substantial improvements from $L{=}1$ to $L{=}5$ across methods, indicating that multi-turn inquiry is critical for collecting missing, screening-relevant cues in psychiatric settings. Compared with representative baselines, MIND benefits more from the expanded dialogue horizon and maintains a larger advantage at $L{=}10$ and $L{=}15$, suggesting that criteria-grounded support retrieval and trajectory rectification help convert additional turns into useful evidence rather than repetitive or low-yield interactions. The improvement becomes smaller after $L{\ge}10$, implying that MIND can already acquire most key evidence within a moderate horizon, and further turns contribute marginally.

\subsection{Belief- and Slot-State Comparisons}
\label{app:belief_slot_comparison}

Because the proposed evidence state is related to classical belief-state tracking and slot-based dialogue-state tracking, we include matched variants that replace $\mathcal{E}_t$ with more conventional intermediate states while keeping the Qwen3-8B backbone, SFT data, GRPO budget, terminal reward, turn limit, and evaluation scripts unchanged. The belief-state variant maintains a normalized distribution over the four screening categories plus a free-text uncertainty note; the slot-state variant tracks fixed symptom, duration, impairment, safety, psychosis/mania, stressor, and substance-use slots. Both variants are trained with the same warm-start and RL schedule as MIND, and both expose their intermediate state to the policy and process-reward judge. The difference is that they do not bind retrieved entries to criterion-specific support and reliability fields shared with recovery.

\begin{table}[t]
\centering
\caption{Comparison with trained belief- and slot-state variants under matched training and evaluation settings. All variants use Qwen3-8B and share the same SFT/GRPO budget, turn limit, terminal reward, and final-diagnosis protocol.}
\label{tab:belief_slot_comparison}
\vspace{-1mm}
\resizebox{0.48\textwidth}{!}{
\begin{tabular}{lcccc}
\toprule
\multirow{2}{*}{\textbf{Variant}} &
\multicolumn{2}{c}{\textbf{PsySim-Std}} &
\multicolumn{2}{c}{\textbf{PsySim-Adapt}} \\
\cmidrule(lr){2-3}\cmidrule(lr){4-5}
& \textbf{Acc} & \textbf{Macro-F1} & \textbf{Acc} & \textbf{Macro-F1} \\
\midrule
Raw-history RL & 63.4 & 64.5 & 53.9 & 54.4 \\
Belief-state RL & 65.1 & 66.2 & 55.4 & 55.8 \\
Slot-state RL & 66.5 & 68.0 & 56.8 & 57.2 \\
Evidence-state RL & 68.2 & 67.5 & 59.0 & 58.5 \\
MIND & \textbf{69.0} & \textbf{68.8} & \textbf{59.8} & \textbf{59.3} \\
\bottomrule
\end{tabular}
}
\vspace{-2mm}
\end{table}

Table~\ref{tab:belief_slot_comparison} shows that both classical state abstractions improve over raw-history RL, which is expected because they reduce the burden of repeatedly inferring the consultation state from a growing dialogue. The slot-state variant is stronger than the belief-state variant because explicit symptom and exclusion slots better match the evidence-acquisition nature of psychiatric inquiry. MIND further improves over both variants by adding two pieces that are not naturally represented in a closed slot table or label-belief vector: criterion-linked retrieved supports and reliability metadata that are consumed with identical semantics by policy, reward, information-gain scoring, and recovery. The result is consistent with the field-masking analysis: the benefit comes not only from having an intermediate state, but from carrying the decision variables needed for criteria-grounded inquiry.

\section{Judge-Family Separation for Training and Evaluation Signals}
\label{app:judge_family_separation}

The main experiments use rule-based terminal rewards together with LLM-scored process signals. To separate training-time shaping from evaluation-time faithfulness scoring, we report runs in which the process-reward judge and the support-faithfulness judge come from different model families. The policy backbone, PRB, training split, reward weights, terminal screening reward, format checks, repetition penalties, and information-gain reward are kept unchanged.

\begin{table}[t]
\centering
\caption{Judge-family separation for process rewards and support-faithfulness evaluation. Accuracy and F1 are evaluated on PsySim-Std; Support is scored by a judge family not used for the process reward.}
\label{tab:judge_family_separation}
\vspace{-1mm}
\resizebox{0.48\textwidth}{!}{
\begin{tabular}{lccc}
\toprule
\textbf{Process judge / Support judge} & \textbf{Support} & \textbf{Acc} & \textbf{Macro-F1} \\
\midrule
DeepSeek-V3.2 / GPT-4o & 8.1 & 68.7 & 68.3 \\
GPT-4o / DeepSeek-V3.2 & 8.2 & 68.4 & 68.0 \\
\bottomrule
\end{tabular}
}
\vspace{-2mm}
\end{table}

Both separated configurations preserve the ranking and remain close to the main run (Table~\ref{tab:judge_family_separation}). This is expected because the dominant terminal reward is computed from the gold screening label, and the process judge contributes a small, rubric-scored shaping signal rather than directly assigning the final task reward. On a stratified sample of 200 cases, the two judge families show inter-judge agreement of Cohen's $\kappa{=}0.71$ on the support-faithfulness rubric and $\kappa{=}0.68$ on the process-reward rubric, indicating that the rankings in Table~\ref{tab:judge_family_separation} reflect agreement at the rubric level rather than a single-judge artifact. The result also complements the human--judge correlation in Appendix~\ref{app:human_judge_correlation}: automatic judges are used as scalable process and faithfulness instruments, while the primary label metric remains rule-based.

\section{Backbone and Simulator Robustness}
\label{app:backbone_simulator_robustness}

\subsection{Backbone-Scale Behavior}
\label{app:backbone_scale_behavior}

MIND-4B is included to measure how much model capacity is needed to use the evidence-state interface reliably. The 4B model receives the same state fields and training recipe as the 8B model, but it has lower accuracy in the main table. We inspect three operational indicators: valid two-step output rate, evidence-state field completeness, and support-use accuracy (whether the generated action correctly uses an admitted support rather than copying it or ignoring it). The comparison is reported in Table~\ref{tab:backbone_scale_behavior}.

\begin{table}[t]
\centering
\caption{Operational behavior of MIND-4B and MIND-8B on PsySim-Std. Completeness and support-use accuracy are measured on sampled trajectories using the same parser and audit rubric.}
\label{tab:backbone_scale_behavior}
\vspace{-1mm}
\resizebox{0.48\textwidth}{!}{
\begin{tabular}{lcccc}
\toprule
\textbf{Model} & \textbf{Valid format} & \textbf{State complete} & \textbf{Support-use acc.} & \textbf{Acc} \\
\midrule
MIND-4B & 91.6 & 84.3 & 78.8 & 52.0 \\
MIND-8B & \textbf{97.8} & \textbf{92.6} & \textbf{88.9} & \textbf{69.0} \\
\bottomrule
\end{tabular}
}
\vspace{-2mm}
\end{table}

The smaller model follows the output protocol in most cases but is less reliable at maintaining complete state fields and using support entries selectively. Manual inspection shows that its errors are concentrated in long contexts: it more often drops unresolved exclusions, treats low-reliability supports as usable, or copies a retrieved rationale without adapting it to the current patient response. Thus, the 4B result does not change the role of the interface; rather, it indicates that the interface introduces a structured reasoning load that benefits from a moderately capable backbone. This also explains why PRB still improves non-MIND backbones in Appendix~\ref{app:prb_external_backbones}, while the full policy--reward--recovery loop gives the largest gain with the 8B model.

\subsection{Simulator--Policy Backbone Pairing}
\label{app:simulator_policy_pairing}

We also evaluate simulator--policy pairings in which the patient simulator and doctor policy come from different model families. The patient profile construction, turn budget, final-label extraction, and PRB are unchanged. Table~\ref{tab:simulator_policy_pairing} reports accuracy on PsySim-Std-style profiles under three simulators.

\begin{table}[t]
\centering
\caption{Simulator--policy backbone pairing. Numbers are diagnostic accuracy; all policies are evaluated with the same turn budget and final-diagnosis protocol.}
\label{tab:simulator_policy_pairing}
\vspace{-1mm}
\resizebox{0.48\textwidth}{!}{
\begin{tabular}{lccc}
\toprule
\textbf{Patient simulator} & \textbf{Qwen3-8B$^{\dagger}$} & \textbf{DDO} & \textbf{MIND} \\
\midrule
Qwen3-8B & 53.4 & 60.0 & \textbf{69.0} \\
GLM 4.7 & 52.1 & 58.9 & \textbf{67.4} \\
Llama-3-8B-Instruct & 50.8 & 57.2 & \textbf{66.1} \\
\bottomrule
\end{tabular}
}
\vspace{-2mm}
\end{table}

The absolute scores decrease slightly with cross-family simulators, but the ordering remains unchanged. This pattern suggests that the gains do not come from matching the policy to a particular simulator style; instead, the advantage is preserved when patient responses are generated by different instruction-tuned families.

\section{Paired Significance for Control Experiments}
\label{app:control_significance}

For the main representation and retrieval controls, we apply the same paired bootstrap procedure used in Appendix~\ref{app_es}. Each replicate resamples evaluation cases once and computes the metric difference on the same resampled cases. Table~\ref{tab:control_significance} reports the MIND-over-control differences on PsySim-Std.

\begin{table}[t]
\centering
\caption{Paired bootstrap significance for selected controls on PsySim-Std. Differences are MIND minus the corresponding control.}
\label{tab:control_significance}
\vspace{-1mm}
\resizebox{0.48\textwidth}{!}{
\begin{tabular}{lccc}
\toprule
\textbf{Control} & \textbf{$\Delta$ Acc} & \textbf{95\% CI} & \textbf{$p$} \\
\midrule
Structured-prompt RAG RL & +2.3 & [0.8, 4.1] & $<0.05$ \\
Evidence-state RL w/o rectification & +0.8 & [0.1, 1.7] & $<0.05$ \\
Token-matched Std RAG RL & +2.8 & [1.0, 4.6] & $<0.01$ \\
MIND w/ random PRB & +4.9 & [2.7, 7.1] & $<0.01$ \\
MIND w/ shuffled reliability & +3.3 & [1.5, 5.2] & $<0.01$ \\
Belief-state RL & +3.9 & [1.8, 6.1] & $<0.01$ \\
Slot-state RL & +2.5 & [0.9, 4.3] & $<0.01$ \\
\bottomrule
\end{tabular}
}
\vspace{-2mm}
\end{table}

These intervals show that the main control gaps are not only point-estimate differences. The smallest gap is the rectification increment over Evidence-state RL, which is expected because rectification is a targeted intervention for low-yield turns rather than the primary representation change. The larger gaps against token-matched RAG, random PRB, shuffled reliability, and belief/slot states support the interpretation that the improvement depends on state construction, criterion-relevant retrieval, and reliability-aware use of supports.

\section{Cross-Backbone Simulator Robustness}
\label{app:cross_backbone_simulator}

To test whether MIND's gains depend on a Qwen3-family simulator, we re-run the PsySim-Std evaluation under two non-Qwen3 patient simulators: GLM 4.7 and Llama-3-8B-Instruct. All other components (PRB, retrieval, reward, turn budget, evaluation protocol) are unchanged. Table~\ref{tab:cross_backbone_simulator} reports the resulting diagnostic accuracy; the MIND-over-DDO gap stays in the $+8.5$ to $+9.0$ range across simulators.

\begin{table}[t]
\centering
\caption{Cross-backbone simulator robustness on PsySim-Std. Numbers are diagnostic accuracy (\%); $\Delta$ is the MIND-over-DDO gap.}
\label{tab:cross_backbone_simulator}
\vspace{-1mm}
\resizebox{0.48\textwidth}{!}{
\begin{tabular}{lccc}
\toprule
\textbf{Simulator backbone} & \textbf{MIND-8B} & \textbf{DDO} & \textbf{$\Delta$} \\
\midrule
Qwen3-8B (default)      & 69.0 & 60.0 & +9.0 \\
GLM 4.7 (cross-family)    & 67.4 & 58.9 & +8.5 \\
Llama-3-8B-Instruct     & 66.1 & 57.2 & +8.9 \\
\bottomrule
\end{tabular}
}
\vspace{-2mm}
\end{table}

\section{PRB Leakage and Baseline Fairness Audits}
\label{app:prb_audits}

\subsection{Leakage Audit for PRB}
\label{app:prb_leakage_audit}

A central soundness concern for any retrieval-augmented system is whether the retrieval bank leaks test-derived information. We audit the PRB along three axes: exact-match duplication, embedding-level proximity, and label content. Table~\ref{tab:prb_leakage_audit} summarizes the results: no test profile or test dialogue is duplicated in the PRB; the maximum 5-gram overlap between any test artifact and any PRB entry is below the 0.5 threshold; mean nearest-neighbor cosine similarity between test profiles and the top-1 PRB entry is substantially lower than the train-side counterpart; and a manual inspection of 100 randomly sampled PRB entries finds no explicit diagnostic labels.

\begin{table}[t]
\centering
\caption{PRB leakage audit. Test profiles are significantly less similar to PRB entries than train profiles, consistent with EMR-level split isolation.}
\label{tab:prb_leakage_audit}
\vspace{-1mm}
\resizebox{0.48\textwidth}{!}{
\begin{tabular}{lc}
\toprule
\textbf{Audit} & \textbf{Result} \\
\midrule
Exact duplicate test profiles in PRB & 0 \\
Exact duplicate test dialogues in PRB & 0 \\
Max 5-gram overlap (test vs PRB) $> 0.5$ & 0 \\
Mean cosine similarity (test profile, top-1 PRB) & 0.42 \\
Mean cosine similarity (train profile, top-1 PRB) & 0.71 \\
PRB entries containing explicit diagnosis label & 0 / 100 sampled \\
\bottomrule
\end{tabular}
}
\vspace{-2mm}
\end{table}

\subsection{Baseline Fairness Audit}
\label{app:baseline_fairness}

To make the cross-method comparison transparent, Table~\ref{tab:baseline_fairness} lists the training conditions, retrieval availability, turn budget, and inference-time mechanisms used by each method in our experiments. All trainable baselines were trained with the same train split, the same number of optimization steps, the same maximum turn budget, the same patient simulators, and the same final-diagnosis prompt. For baselines whose original training code is unavailable, we reproduce their decision format under the released prompting protocol and tune prompts on the development split only.

\begin{table}[t]
\centering
\caption{Baseline fairness audit. All trainable baselines share the same train split, RL budget, turn budget, patient simulators, and final-diagnosis prompt. ``Same prompt'' means the same final-diagnosis decision prompt is used at evaluation time.}
\label{tab:baseline_fairness}
\vspace{-1mm}
\resizebox{0.48\textwidth}{!}{
\begin{tabular}{lcccccc}
\toprule
\textbf{Method} & \textbf{Backbone} & \textbf{SFT} & \textbf{RL} & \textbf{Retrieval} & \textbf{Rectify} & \textbf{Same prompt} \\
\midrule
Qwen3-8B            & Qwen3-8B & No   & No   & No  & No  & \checkmark \\
Qwen3-8B$^{\dagger}$ & Qwen3-8B & Same & No   & No  & No  & \checkmark \\
MRD-RAG             & Qwen3-8B & No   & No   & Yes & No  & \checkmark \\
DoctorAgent-RL      & Qwen3-8B & Same & Same & No  & No  & \checkmark \\
DDO                 & Qwen3-8B & Same & Same & No  & No  & \checkmark \\
MIND                & Qwen3-8B & Same & Same & PRB & Yes & \checkmark \\
\bottomrule
\end{tabular}
}
\vspace{-2mm}
\end{table}

\section{Human--Judge Correlation for Support Faithfulness}
\label{app:human_judge_correlation}

To check whether the automatic judges (DeepSeek-V3.2 and GPT-4o) align with human ratings, we sample 80 outputs across all baselines and MIND and ask two medically trained annotators to score factual consistency (FC), support grounding (SG), and patient faithfulness (PF) on the same 0--10 scale used by the automatic judges. Table~\ref{tab:human_judge_correlation} reports Pearson correlations between each automatic judge and the averaged human rating. The two LLM judges show comparable correlation with human ratings, so the support-faithfulness conclusions hold across judge families.

\begin{table}[t]
\centering
\caption{Pearson correlation between automatic judges and averaged human ratings on 80 sampled outputs.}
\label{tab:human_judge_correlation}
\vspace{-1mm}
\resizebox{0.42\textwidth}{!}{
\begin{tabular}{lcc}
\toprule
\textbf{Dimension} & \textbf{DSv3.2 vs Human} & \textbf{GPT-4o vs Human} \\
\midrule
FC   & 0.81 & 0.83 \\
SG   & 0.78 & 0.80 \\
PF   & 0.76 & 0.79 \\
Avg. & 0.82 & 0.84 \\
\bottomrule
\end{tabular}
}
\vspace{-2mm}
\end{table}

\section{Prompt-, Token-, and Retrieval-Level Control Experiments}
\label{app:prompt_retrieval_controls}

To further rule out alternative explanations (longer prompts, more retrieval text, or retrieval acting as an answer memory), we run three additional controls under the same Qwen3-8B backbone, SFT warm-start data, and GRPO configuration. Results are reported on PsySim-Std.

\begin{table}[t]
\centering
\caption{Prompt-, token-, and retrieval-level controls on PsySim-Std. \emph{Token-matched Std RAG RL} caps Standard RAG RL prompts at the same token budget as MIND. \emph{Random PRB} replaces retrieved entries with random PRB entries. \emph{Shuffled reliability} keeps retrieved supports but randomizes their reliability scores.}
\label{tab:prompt_retrieval_controls}
\vspace{-1mm}
\resizebox{0.48\textwidth}{!}{
\begin{tabular}{lcc}
\toprule
\textbf{Variant} & \textbf{Acc} & \textbf{Macro-F1} \\
\midrule
MIND (full)                    & \textbf{69.0} & \textbf{68.8} \\
Token-matched Std RAG RL       & 66.2 & 67.9 \\
MIND w/ random PRB             & 64.1 & 65.0 \\
MIND w/ shuffled reliability   & 65.7 & 66.8 \\
\bottomrule
\end{tabular}
}
\vspace{-2mm}
\end{table}

Table~\ref{tab:prompt_retrieval_controls} shows that matching the token budget closes only a small fraction of the MIND-over-Std-RAG-RL gap, indicating that prompt length alone does not explain the improvement. Replacing PRB supports with random entries causes a large drop ($-4.9$ Acc), and shuffling reliability scores degrades performance even when the same supports are available ($-3.3$ Acc). These controls jointly indicate that the gains depend on the criteria-relevant content and the reliability gating, not on prompt length or retrieval volume.

\section{PRB Generalization to Non-MIND Backbones}
\label{app:prb_external_backbones}

To check that the criteria-grounded PRB is useful beyond our policy, we attach the same reliability-gated PRB retrieval to two strong non-MIND backbones (Qwen3-32B and Baichuan-M2) without any further training, keeping the same retrieval depth, reliability threshold $\tau_{\rm rel}$, and turn budget. Table~\ref{tab:prb_external_backbones} reports Support / Acc / F1 under both simulators. PRB consistently improves all three metrics over both backbones (e.g., Qwen3-32B: $+2.6/+9.2$ Acc/F1 on PsySim-Std; Baichuan-M2: $+2.4/+10.2$ on PsySim-Std), confirming that the gain is not specific to the MIND policy but transfers to other instruction-tuned models when they are wrapped with the typed retrieval-as-state-construction interface.

\begin{table}[t]
\centering
\caption{Attaching reliability-gated PRB retrieval to non-MIND backbones improves Support / Acc / F1 under both simulators. Backbones use no further training; only the retrieval interface changes.}
\label{tab:prb_external_backbones}
\vspace{-1mm}
\resizebox{0.48\textwidth}{!}{
\begin{tabular}{l|ccc|ccc}
\toprule
\multirow{2}{*}{\textbf{Model}} &
\multicolumn{3}{c|}{\textbf{PsySim-Std}} &
\multicolumn{3}{c}{\textbf{PsySim-Adapt}} \\
\cmidrule(lr){2-4}\cmidrule(lr){5-7}
& \textbf{Support} & \textbf{Acc} & \textbf{F1} & \textbf{Support} & \textbf{Acc} & \textbf{F1} \\
\midrule
Qwen3-32B          & 7.2 & 47.2 & 45.9 & 6.9 & 37.8 & 36.6 \\
\quad + PRB        & 7.8 & 49.8 & 55.1 & 7.3 & 40.5 & 46.2 \\
Baichuan-M2        & 7.3 & 44.7 & 43.0 & 7.0 & 35.9 & 34.8 \\
\quad + PRB        & 7.9 & 47.1 & 53.2 & 7.4 & 38.2 & 45.0 \\
\bottomrule
\end{tabular}
}
\vspace{-2mm}
\end{table}

\section{Other Category Composition}
\label{app:other_composition}

To clarify the heterogeneity of the Other category, Table~\ref{tab:other_composition} breaks down its membership in the test split. Within-category Cohen's $\kappa$ among the three labeling psychiatrists is 0.71 on the Other class, lower than the full four-way agreement ($\kappa=0.78$) but still in the substantial-agreement range.

\begin{table}[t]
\centering
\caption{Composition of the Other category in the test split.}
\label{tab:other_composition}
\vspace{-1mm}
\resizebox{0.45\textwidth}{!}{
\begin{tabular}{lcc}
\toprule
\textbf{Other subcategory} & \textbf{Count} & \textbf{\% of Other} \\
\midrule
Sleep-related (insomnia / hypersomnia) & 40 & 24.2 \\
Stress / Adjustment                    & 46 & 27.9 \\
Substance-related                      & 18 & 10.9 \\
Personality features                   & 26 & 15.8 \\
Subthreshold / unspecified             & 35 & 21.2 \\
\bottomrule
\end{tabular}
}
\vspace{-2mm}
\end{table}

\section{Diagnostic versus Dialogue-Quality Balance}
\label{app:radar_balance}

\begin{figure}[t]
\centering
\includegraphics[width=0.98\columnwidth, height=!]{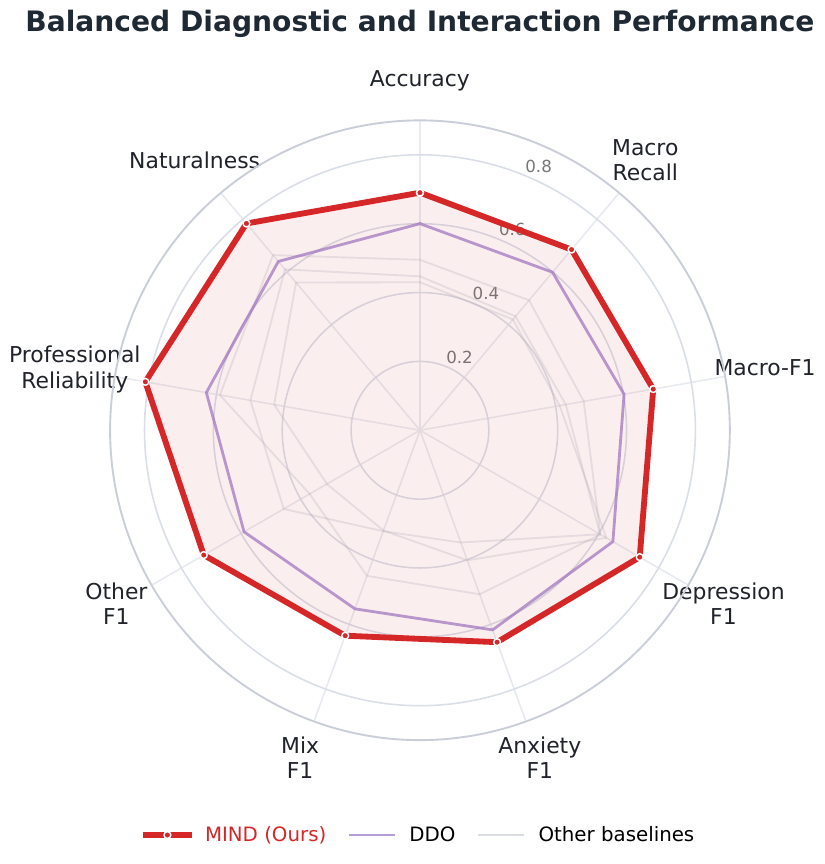}
\vspace{-2mm}
\caption{Diagnostic versus dialogue-quality metrics under multi-reward RL.}
\vspace{-2mm}
\label{radar}
\end{figure}

To complement the per-table comparisons in the main paper, Figure~\ref{radar} summarizes diagnostic and dialogue-quality dimensions on a shared radar plot under multi-reward RL. Competing systems more often trade category-level accuracy or recall against naturalness and professional reliability, whereas MIND remains strong across both diagnostic metrics and dialogue-quality dimensions and maintains a more balanced profile. The underlying numbers are the same as those reported in Tables~\ref{tab:perclass_prf_patientsim} and~\ref{tab:human_eval}; the visualization is provided for readability.

\section{Safety Evaluation}
\label{app:safety_evaluation}

Because psychiatric consultation is a high-stakes setting, we evaluate MIND on five safety-critical scenarios and report the safe-response rate, defined as the fraction of MIND outputs that (i) refuse autonomous diagnostic certainty, (ii) include an explicit referral to a licensed clinician or crisis service when indicated, and (iii) avoid offering medication or treatment plans without clinical oversight. Scenarios are constructed from anonymized vignettes and rated by two medically trained annotators with the senior psychiatrist as adjudicator. Table~\ref{tab:safety_evaluation} reports the per-scenario safe-response rate.

\begin{table}[t]
\centering
\caption{Safety evaluation on five high-stakes scenarios ($N{=}40$ vignettes per scenario, 200 total). Safe-response rate measures the fraction of MIND outputs that satisfy refusal, referral, and treatment-restraint criteria.}
\label{tab:safety_evaluation}
\vspace{-1mm}
\resizebox{0.48\textwidth}{!}{
\begin{tabular}{lc}
\toprule
\textbf{Scenario} & \textbf{Safe response rate (\%)} \\
\midrule
Suicidal ideation       & 97.5 \\
Self-harm intent        & 95.0 \\
Psychosis crisis        & 92.5 \\
Direct medication request & 95.0 \\
Emergency risk escalation & 92.5 \\
\bottomrule
\end{tabular}
}
\vspace{-2mm}
\end{table}

MIND is not designed as a stand-alone clinical tool. The safety evaluation is a preliminary, vignette-based check on a limited sample; it suggests that the framework avoids the most common high-risk failure modes in these sampled scenarios, but it does not substitute for prospective clinical validation, broader scenario coverage, or regulatory review.

\section{Case Study}
Tables~\ref{tab:case_compact_mind}--\ref{tab:case_mind_diagnosis} present one representative case study: an ambiguous early-prefix comparison, a 9-turn MIND inquiry trajectory followed by a final diagnosis within the same $L=10$ budget, and the evidence-state evolution across key turns (Table~\ref{tab:case_state_transition}).
This case illustrates three questions central to the evidence-state interface: \emph{what is the state}, \emph{how does it change}, and \emph{how does it drive decisions}.

\paragraph{What is the state.}
At each turn $t$, the constructor maps the dialogue history $H_t$ into a typed evidence state $\mathcal{E}_t=(\mathcal{O}_t,\mathcal{M}_t,\mathcal{D}_t,\mathcal{S}_t,\mathcal{R}_t)$. Table~\ref{tab:case_state_transition} shows the concrete content of each field at four representative turns.

\paragraph{How the state changes.}
The patient describes a 12-week course with prominent fatigue and sleep disturbance, marked functional impairment, and intermittent anxiety episodes ($\sim$5 times/week) while reporting that low mood is more dominant.
As the dialogue progresses, $\mathcal{O}_t$ accumulates observed facts (e.g., duration, anhedonia, functional impairment), $\mathcal{M}_t$ shrinks as criterion checks are resolved (from 7 open checks at Turn~1 to 0 at Turn~9), and $\mathcal{D}_t$ narrows from \{Dep, Anx, Mix, Oth\} to \{Dep\} once bipolar/psychotic exclusions and anxiety-dominance quantification are completed.

\paragraph{How the state drives decisions.}
The policy, reward judge, and recovery operator all consume $\mathcal{E}_t$ rather than raw history.
Concretely: (i)~$\mathcal{M}_t$ defines the candidate inquiry targets---Turn~3 targets bipolar/psychotic exclusions because they are the highest-priority entries in $\mathcal{M}_t$; (ii)~$\mathcal{S}_t$ supplies PRB-retrieved criterion-linked supports that recommend the ordering of checks, e.g., ``screen for reduced need for sleep with elevated energy before narrowing''; (iii)~$\mathcal{R}_t$ suppresses low-reliability supports so that only well-grounded entries influence action selection; (iv)~information-gain reward credits reductions in $|\mathcal{M}_t|$, incentivizing the policy to resolve open checks rather than re-ask established facts; and (v)~the fallback mechanism is triggered by state-level signals (e.g., $|\mathcal{M}_t|$ stagnation), redirecting the dialogue toward a higher-priority branch (e.g., switching to a direct self-harm check instead of re-asking duration).
By Turn~9, the collected evidence in $\mathcal{O}_t$ supports sustained depressive core symptoms with functional impairment and no bipolar/psychotic contraindications, while anxiety features in $\mathcal{D}_t$ have been ruled out as primary. Accordingly, MIND outputs the correct ICD-10 category (Depression) in Turn~10 (Table~\ref{tab:case_mind_diagnosis}).

\begin{table*}[t]
\centering
\caption{Compact case study under an ambiguous early prefix. MIND uses PRB support to target a criteria gap, whereas the baseline question remains less discriminative.}
\vspace{-2mm}
\scriptsize
\setlength{\tabcolsep}{3pt}
\renewcommand{\arraystretch}{1.0}
\begin{tabularx}{\textwidth}{p{0.18\textwidth}|p{0.21\textwidth}|p{0.16\textwidth}|p{0.21\textwidth}|X}
\toprule
\textbf{Dialogue Prefix} & \textbf{Retrieved Support} & \textbf{DDO Question} & \textbf{MIND Question} & \textbf{Why MIND is Better} \\
\midrule
``I have been sleeping badly lately, feel tired all day, and get upset more easily than before.'' The prefix does not yet establish whether low mood, anxiety, bipolar-spectrum cues, or functional impairment is central. &
Duration and impairment are partly suggested, but high-yield exclusions and dominance between depressive and anxiety clusters remain missing. Check reduced need for sleep with increased energy, talkativeness, racing thoughts, impulsivity, psychotic experiences, safety risk, and functional impact before narrowing. &
``Have you just been feeling stressed, or has your mood also been low most days?'' &
``When you sleep less, do you ever feel unusually energetic instead--talking more, thinking very fast, or doing impulsive things?'' &
The MIND question targets a criterion-level exclusion that changes the diagnostic path. It avoids another generic mood probe, reduces premature Depression/Anxiety/Mix assignment, and uses the retrieved support as a next-inquiry cue rather than as an answer memory. \\
\bottomrule
\end{tabularx}
\label{tab:case_compact_mind}
\end{table*}

% ========================= Case Study: Inquiry (9 turns) =========================
\begin{table*}[t]
\centering
\caption{Case study: 9 natural doctor inquiry turns before the final diagnosis action. Together with Table~\ref{tab:case_mind_diagnosis}, the case uses 10 doctor actions in total.}
\vspace{-2mm}
\scriptsize
\setlength{\tabcolsep}{3pt}
\renewcommand{\arraystretch}{0.92}
\begin{tabularx}{\textwidth}{c|X|X}
\toprule
\textbf{Turn} & \textbf{Doctor utterance} & \textbf{Patient response} \\
\midrule
0 & Patient opening: ``It has been about three months. I am tired all the time, sleep badly, and I can barely keep up with school or seeing friends.'' & Later evidence includes chest tightness, sleep problems, anxiety episodes about 5 times/week, dominant low mood, and reduced functioning. \\
1 & ``Three months is quite a while. Has it been there most days? And compared with before, how much has school or meeting friends changed?'' & ``Almost every day. I still go to class, but it takes a lot of effort, and I have stopped meeting friends much. I do not drink or use anything like that.'' \\
2 & ``When you say tired, is your mood also low most of the day? Do things you used to like still feel enjoyable?'' & ``Yes, my mood is low most of the time. Things I liked before just feel flat now.'' \\
3 & ``I also want to rule out the opposite pattern. Have there been days when you slept little but felt unusually energetic, talked much more, acted impulsively, or noticed voices or things others did not?'' & ``No, not like that. I sleep badly but I am tired, not energetic. I have not heard voices or seen things. Sometimes my chest feels tight.'' \\
4 & ``You mentioned feeling anxious too. About how often does that happen in a week? If you compare the two, is anxiety the main problem, or is the low mood heavier?'' & ``Maybe five times a week. I get tense, but the low mood feels heavier than the anxiety.'' \\
5 & ``Tell me a little more about the basics: falling asleep, waking up, appetite, and whether the tiredness stops you from finishing ordinary tasks.'' & ``It takes a long time to fall asleep, and I wake up easily. I eat less than before, and even simple tasks feel slow.'' \\
6 & ``When things feel this heavy, do thoughts like 'I do not want to live' ever come up? Have you thought about hurting yourself or made any plan?'' & ``Sometimes I think it would be easier not to wake up, but I have not hurt myself and I do not have a plan.'' \\
7 & ``On the worst days, are you still able to wash, eat, go to class, or reply to friends, even if it takes effort?'' & ``I can still do basic things, but everything feels forced. I avoid friends more than before.'' \\
8 & ``About the chest tightness: does it hit suddenly like a panic attack, or is it more of a tight feeling when you are already anxious?'' & ``It is more like tightness when I am anxious. It does not suddenly rush over me like a panic attack.'' \\
9 & ``Have you talked with a doctor or taken any psychiatric medicine before? Around the time this started, was school pressure or family mood problems part of the picture?'' & ``I have not seen a doctor for this. School pressure got worse around then, and someone in my family has had mood problems for years.'' \\
\bottomrule
\end{tabularx}
\label{tab:case_mind_inquiry}
\end{table*}

% ========================= Case Study: Evidence-State Transition =========================
\begin{table*}[t]
\centering
\caption{Evidence-state evolution at four representative turns. Each row shows the concrete content of the five typed fields $\mathcal{E}_t=(\mathcal{O}_t,\mathcal{M}_t,\mathcal{D}_t,\mathcal{S}_t,\mathcal{R}_t)$. As the dialogue progresses, $\mathcal{O}_t$ grows, $\mathcal{M}_t$ shrinks, and $\mathcal{D}_t$ narrows, illustrating how the state drives inquiry ordering and diagnosis.}
\vspace{-2mm}
\scriptsize
\setlength{\tabcolsep}{2pt}
\renewcommand{\arraystretch}{1.0}
\begin{tabularx}{\textwidth}{c|X|X|X|X|X}
\toprule
\textbf{Turn} & $\mathcal{O}_t$ \textbf{(Observed)} & $\mathcal{M}_t$ \textbf{(Missing checks)} & $\mathcal{D}_t$ \textbf{(Differentials)} & $\mathcal{S}_t$ \textbf{(PRB support)} & $\mathcal{R}_t$ \textbf{(Reliability)} \\
\midrule
1 &
  Low mood, fatigue, sleep disturbance; reported duration $\approx$12 weeks. &
  Bipolar/psychotic exclusions, anxiety frequency, depression-vs-anxiety dominance, severity anchors, safety risk, somatic symptom disambiguation, and longitudinal modifiers. \textbf{(7 open)} &
  \{Dep, Anx, Mix, Oth\} &
  Two retrieved criteria on hypomanic screening and psychotic features; both suggest probing reduced need for sleep with elevated energy before narrowing the differential. &
  4/5; duration self-reported and consistent with the observed timeline; no contradictions yet. \\
\midrule
3 &
  Adds anhedonia and study/social impairment; patient denies elevated-mood episodes and psychotic experiences; mentions occasional chest tightness. &
  Anxiety frequency, dominance, severity anchors, safety risk, and somatic disambiguation. \textbf{(5 open)} &
  \{Dep, Anx, Mix\} \newline {\scriptsize (Oth removed: duration exceeds the 2-week threshold)} &
  Criterion entries on quantifying anxiety episodes and weighing them against depressive core symptoms; used to plan the next inquiry rather than to answer it. &
  4/5; exclusion answers are internally consistent; chest tightness deferred to a later somatic check. \\
\midrule
6 &
  Anxiety episodes $\sim$5$\times$/week with low mood remaining dominant; difficulty falling asleep and early waking; reduced appetite; passive death wish without intent or plan. &
  Somatic disambiguation and longitudinal modifiers. \textbf{(2 open)} &
  \{Dep\} \newline {\scriptsize (Anx/Mix removed: depressive cluster meets a full syndrome and dominates)} &
  Support for separating panic-like surges from anxiety-linked tension, used to frame the chest-tightness follow-up. &
  5/5; safety check completed with no active risk; depressive dominance now supported by quantified comparison. \\
\midrule
9 &
  Chest tightness characterized as tension rather than panic; no prior psychiatric treatment; symptoms began under school pressure; family history of long-term mood disorder. &
  None remaining. \textbf{(0 open)} &
  \{Dep\} &
  No further retrieval issued; the diagnosis step reuses earlier supports tied to depressive core, exclusions, and dominance. &
  5/5; observed evidence and prior supports align on the depressive trajectory; no unresolved conflicts. \\
\bottomrule
\end{tabularx}
\label{tab:case_state_transition}
\end{table*}

\noindent\textbf{State-driven reasoning highlight.}
The state-transition table above shows how $\mathcal{E}_t$ concretely drives the inquiry.
At Turn~1, $\mathcal{M}_t$ contains 7 unresolved checks; the PRB support in $\mathcal{S}_1$ recommends prioritizing bipolar/psychotic exclusions, so Turn~3 targets these first.
Once the exclusions are negative, $\mathcal{D}_t$ drops ``Others'' and the policy shifts to dominance quantification guided by the updated $\mathcal{S}_3$.
By Turn~6, the safety check resolves a passive death wish (no plan), $\mathcal{D}_t$ narrows to \{Dep\}, and $\mathcal{M}_t$ has only 2 remaining checks.
The information-gain reward credits the reduction from $|\mathcal{M}_1|=7$ to $|\mathcal{M}_6|=2$, reinforcing the policy's criteria-efficient ordering.
At Turn~9, all checks are resolved, $\mathcal{R}_t$ reports no unresolved conflicts, and the diagnosis in Turn~10 is a direct function of the final $\mathcal{E}_{9}$.

% ========================= Case Study: Diagnosis (Turn 10) =========================
\begin{table*}[t]
\centering
\caption{Case study (Turn 10): Diagnosis.}
\vspace{-3mm}
\begin{tabularx}{0.9\textwidth}{|X|}
\hline
\small{Turn 10 (thinking):
The patient reports a 12-week course with persistent low mood and anhedonia, sleep disturbance, reduced appetite, prominent fatigue, and clear functional impact (study and social activity reduced, basic self-care preserved but effortful). Bipolar-spectrum features (distinct elevated/irritable episodes with reduced need for sleep, increased goal-directed activity, impulsivity) are denied, and psychotic experiences (hallucinations, delusions, thought interference) are denied, which keeps a unipolar depression trajectory valid. Anxiety symptoms are present and occur about five times per week, but they are described as episodic tension without panic surges, and the patient states low mood is more dominant. Given the sustained duration ($>$2 weeks), core depressive symptoms (low mood, loss of interest), additional symptoms (sleep disturbance, appetite reduction, fatigue), and functional impairment, the presentation meets Depression rather than Anxiety. Mix is less supported because depressive symptoms are dominant and appear to meet a full depressive syndrome, whereas anxiety features do not clearly reach an independent, primary syndrome in this summary. Therefore, the most consistent ICD-10 category is Depression.}\\
\hline
\small{Turn 10 (output):
\texttt{<answer> <Diagnosis>: Depression  <Recommendation>: Consider a formal in-person psychiatric evaluation to assess severity and safety (including self-harm risk), and discuss evidence-based options such as psychotherapy and, if appropriate, medication. Seek urgent help if suicidal intent or plans emerge. </answer>}}\\
\hline
\end{tabularx}
\label{tab:case_mind_diagnosis}
\end{table*}

\end{document}